\documentclass[dvipsnames]{article}

\usepackage[final]{lychee}
\usepackage[utf8]{inputenc} 
\usepackage[T1]{fontenc}    
\usepackage{url}            
\usepackage{booktabs}       
\usepackage{amsfonts}       
\usepackage{nicefrac}       
\usepackage{booktabs} 
\usepackage{multirow}
\usepackage{wrapfig}
\usepackage{amssymb}
\usepackage{epigraph}
\usepackage{makecell}
\usepackage{listings}
\usepackage{subcaption}
\usepackage[labelfont=bf]{caption} 

\usepackage[ruled,vlined]{algorithm2e}
\usepackage{color, soul}
\usepackage{microtype}      
\usepackage{xcolor}         
\usepackage{geometry}
\usepackage{times}
\usepackage{ragged2e}
\usepackage{amsmath}
\usepackage{bm}
\usepackage{latexsym}
\usepackage{longtable}
\usepackage{graphicx}
\usepackage{float}
\usepackage{placeins}
\usepackage{booktabs} 
\usepackage{multirow}
\usepackage{amssymb}
\usepackage{tabu}
\usepackage{bbding}
\usepackage{awesomebox}
\usepackage{bbding}
\usepackage[most,breakable]{tcolorbox}
\usepackage{etoolbox}
\usepackage{fancyhdr}
\usepackage{lipsum}
\usepackage{CJKutf8}
\usepackage{pdfpages}
\usepackage{multicol}
\usepackage{pgffor} 
\usepackage{fontawesome5}
\usepackage{nicematrix} 
\usepackage{paralist}
\usepackage[edges]{forest}
\usepackage{siunitx}
\usepackage{tikz-dependency}
\usepackage{tikz}
\usepackage{bbm}
\usepackage{amssymb}
\usepackage[english]{babel}  
\usepackage{hyphenat}
\usepackage{siunitx}

\newcommand{\Rmnum}[1]{\expandafter\@slowromancap\romannumeral #1@}

\usepackage{caption} 
\usepackage{chngcntr}

\usepackage[colorlinks, linkcolor=RoyalBlue, anchorcolor=BrickRed, citecolor=RoyalBlue, urlcolor=RoyalBlue]{hyperref}

\usepackage{cleveref}
\crefname{section}{§}{§§}
\Crefname{section}{§}{§§}

\usepackage{xspace}
\newcommand\ourmodel{\textsc{KnowAct-GUIClaw}\xspace}

\newcommand\refsec[1]{Section~\hyperref[sec:#1]{\ref{sec:#1}}}
\newcommand\refsecs[2]{\hyperref[sec:#1]{§\ref{sec:#1}:~\textsc{#1}}, \hyperref[sec:#2]{§\ref{sec:#2}:~\textsc{#2}}}

\usepackage[tikz]{bclogo}
\definecolor{msftBlue}{RGB}{0,164,239}
\definecolor{msftGreen}{RGB}{127,186,0}
\definecolor{msftYello}{RGB}{255,185,0}
\definecolor{mypurple}{RGB}{138,43,226} 
\definecolor{msftBlack}{RGB}{0,0,0}

\newtcolorbox{myboxnote}[1][]{
  breakable,
  title=#1,
  colback=cyan!0,
  colbacktitle=cyan!0,
  coltitle=black,
  fonttitle=\bfseries,
  bottomrule=0pt,
  toprule=0pt,
  leftrule=1.5pt,
  rightrule=1.5pt,
  titlerule=0pt,
  arc=0pt,
  outer arc=0pt,
  colframe=lightgray,
}
\usepackage{mdframed}

\definecolor{academicblue}{RGB}{54, 95, 145}

\tcbset{
  iclrtakeawaybox/.style={
    width=\linewidth,
    colback=gray!5,                 
    colframe=gray!60,              
    colbacktitle=gray!30,            
    coltitle=black,                
    fonttitle=\bfseries,     
    boxrule=0.5pt,                 
    arc=2pt,                       
    sharp corners=south,           
    attach boxed title to top left={yshift=-0.1in,xshift=0.15in},
    boxed title style={boxrule=0pt, colframe=white},
    enhanced,
    drop shadow southeast           
  }
}
\newtcolorbox{TakeawayBox}[2][]{iclrtakeawaybox,title=#2,#1}

\newenvironment{itemsize*}%
 {\leftmargini=20pt\begin{itemize}%
  \setlength{\itemsep}{3pt}%
  \setlength{\parskip}{0pt}%
  }%
 {\end{itemize}}
\newenvironment{enumerate*}%
 {\begin{enumerate}%
  \setlength{\itemsep}{0pt}%
  \setlength{\parskip}{0pt}}%
 {\end{enumerate}}

\usepackage{fancyhdr} 
\usepackage{blindtext} 
\usepackage{ulem}

\usepackage{xparse}
\usepackage{colortbl}

\title{
\vspace{-2em}
\fontsize{16}{19}\selectfont KnowAct-GUIClaw: Know Deeply, Act Perfectly, \\ Personal GUI Assistant with Self-Evolving Memory and Skill}
\author{
Lychee Team\thanks{Contributors are shown in Sec. \ref{sec:contributors}}, ~Harbin Institute of Technology, Shenzhen \\
AI Training Platform, Shenzhen Loop Area Institute \\[6pt]
\href{https://github.com/HITsz-TMG/KnowAct}{\includegraphics[height=1em]{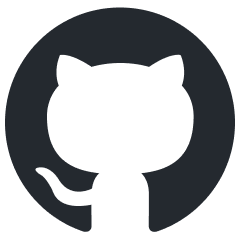} Codes}
\href{https://shibosusu.github.io/KnowAct-GUIClaw/}{\includegraphics[height=1em]{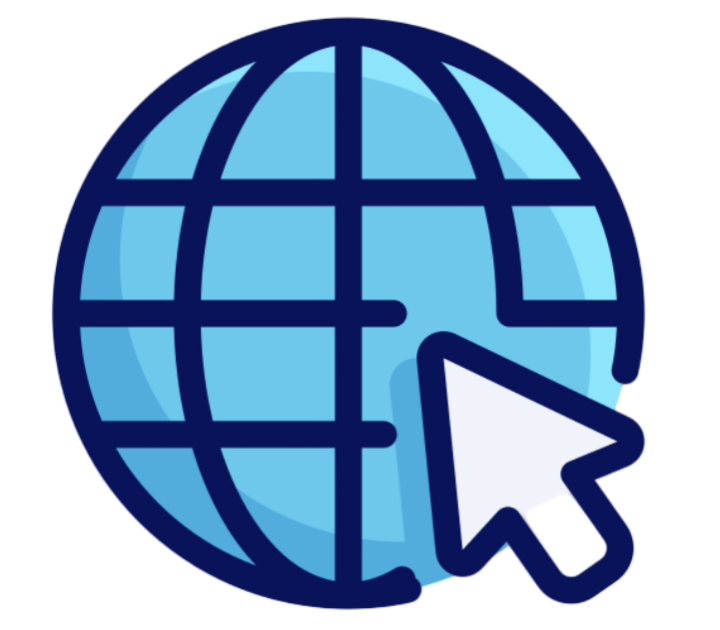} Website}
\href{https://github.com/HITsz-TMG/KnowAct/releases/tag/Result}{\includegraphics[height=1em]{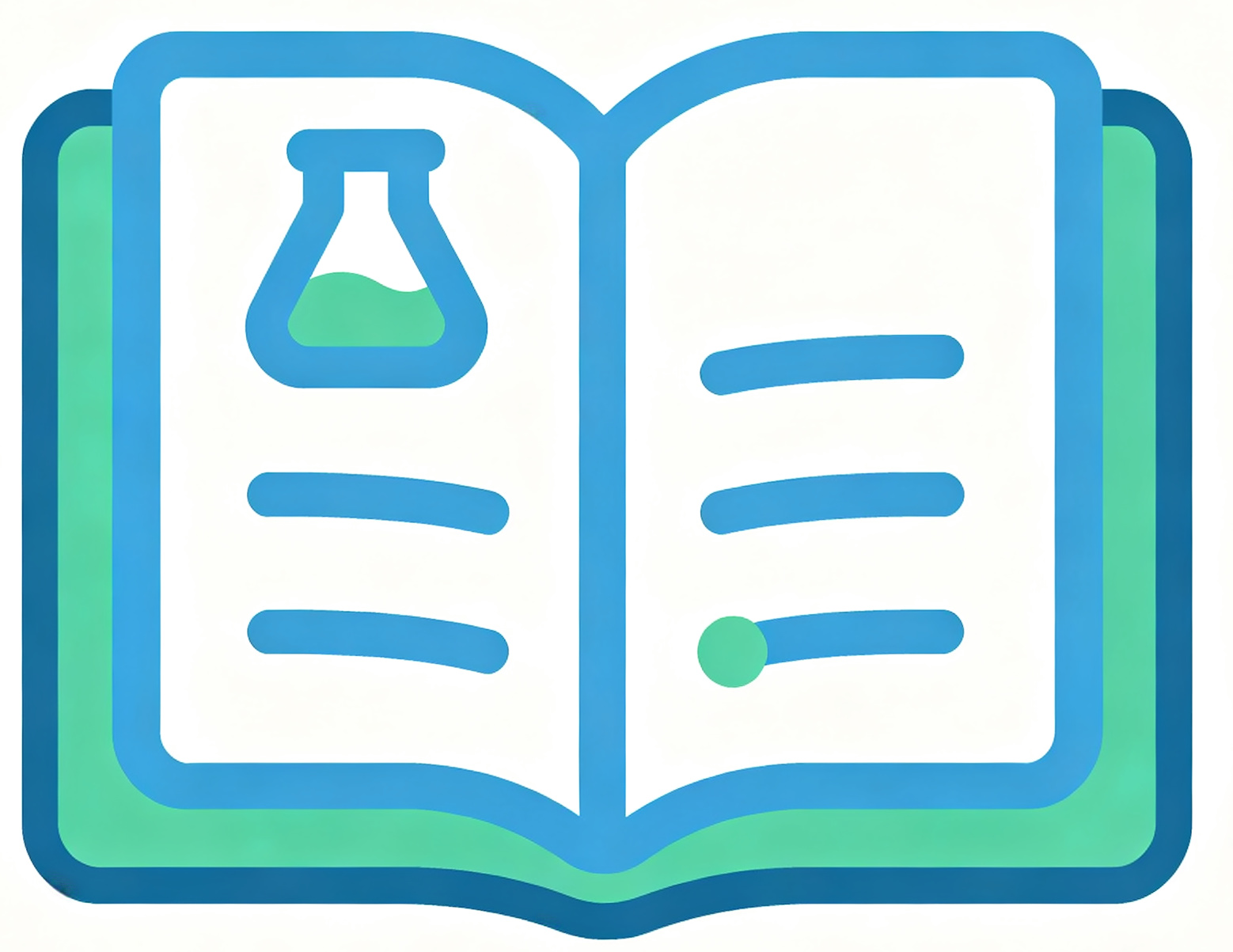} Experimental Logs}
}

\begin{document}

\maketitle
\vspace{-1em}

\begin{abstract}
OpenClaw has emerged as a leading agent framework for complex task automation, yet its existing variants face two core bottlenecks: insufficient cross-platform GUI interaction support and no built-in self-evolution mechanism. These flaws limit its adaptation to heterogeneous device ecosystems and prevent performance improvements through continuous learning from execution experience.
To resolve these issues, we propose the ``\textbf{Know Deeply, Act Perfectly}'' paradigm for personal assistants, which holds that accumulated human-machine interaction and task-running experience directly improve execution accuracy and efficiency, unifying cognitive comprehension and operational execution. Based on this paradigm, we introduce \textbf{KnowAct-GUIClaw}, a novel Know-Route-Act-Reflect framework designed to address OpenClaw’s GUI manipulation deficits and break through its cross-platform and recursive self-improvement constraints. First, the host agent leverages accumulated interaction experience and task-relevant knowledge for long-horizon task decomposition and allocation (Know). Second, a pluggable GUI subagent with an experience-attributable memory system (Know) and self-evolving skill library (Act), enabling seamless cross-platform migration and fast-path integration. Especially, this framework continuously stores user profiles and feedback to improve the accuracy of task decomposition and tool calls.
Extensive experiments across Android, iOS, HarmonyOS and Windows show that \ourmodel achieves superior UI manipulation efficiency, accuracy and cross-platform adaptability. Especially, the GUIClaw with open-source Kimi-2.6 models achieves the best performance (64.1\%) on the long-horizon MobileWorld benchmark, beating all agential frameworks and closed-source agentical models, e.g., Seed-2.0-Pro and GPT-5.5. Additionally, the knowledgeable memory and execution skills supported by our framework are transferable across diverse base models, improving by 8.5\% with Kimi-2.6 and 16.2\% with Qwen3.5-35B-A3B.
\end{abstract}

\begin{figure}[H]
    \centering
    \includegraphics[width=\linewidth]{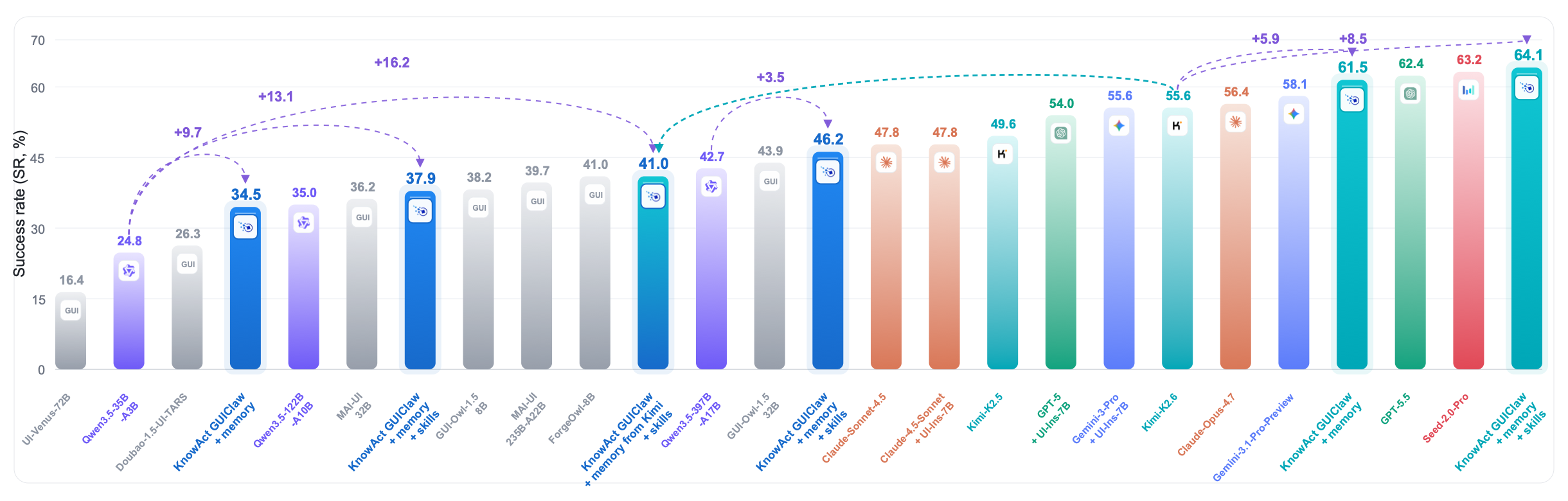}
    \caption{The success rate (SR) comparison on MobileWorld GUI-Only tasks. The bars summarize Table~\ref{tab:mobileworld-leaderboard} together with the additional Kimi-based \ourmodel runs; gray bars denote specialized GUI models, colored external bars denote general model families, and highlighted bars denote \ourmodel variants with memory and skills. The experimental results show that \ourmodel achieves SOTA performance and that the memory and skill are effective for different base models.}
    \label{fig:front}
\end{figure}

\newpage
{
  \hypersetup{linkcolor=RoyalBlue, linktoc=page}
  \tableofcontents
}

\newpage

\section{Introduction}
\label{sec:introduction}

Large language model (LLM) agents are transitioning from one-off dialogue bots to stateful, long-running personal assistants. They process user requests from diverse input channels, preserve persistent workspace state, invoke external tools, coordinate auxiliary subagents, and resume interrupted long-duration workflows. Two dominant architectural lines guide their development: ReAct-based agents unify logical reasoning and environmental action execution~\citep{yao2023reactsynergizingreasoningacting}, while general autonomous agent frameworks modularize planning, memory, tool utilization, and multi-agent dialogue as interchangeable runtime building blocks~\citep{zhou2023agentsopensourceframeworkautonomous}. Practical deployed systems--OpenClaw, Nanobot, and Hermes--integrate these foundational designs into local-first assistant platforms with built-in channel routing, tool extensions, configurable skills, session persistence, and structured memory storage~\citep{openclaw2026,nanobot2026,hermesagent2026}. As capable autonomous personal assistants, these platforms autonomously decompose and fulfil complex user objectives through targeted external tool calls.

However, numerous real-world user tasks demand interaction with graphical user interfaces (GUIs), rather than well-structured, standardized application programming interfaces (APIs). For instance, a personal assistant may be required to inspect mobile applications, migrate data across different apps, handle permission pop-up dialogues, or execute multi-step workflows within login-protected environments. Recent benchmarks spanning web, mobile, and desktop platforms collectively indicate that GUI manipulation poses unique challenges that mandate visual grounding, sequential decision-making, and resilient execution amid dynamically shifting interface states~\citep{deng2023mind2webgeneralistagentweb,zhou2024webarenarealisticwebenvironment,koh2024visualwebarenaevaluatingmultimodalagents,xie2024osworldbenchmarkingmultimodalagents,rawles2025androidworlddynamicbenchmarkingenvironment,li2026windowsworld}. A suite of dedicated GUI agents—including AppAgent, Mobile-Agent, CogAgent, OS-Atlas, ShowUI, UI-TARS, and Aguvis—have delivered remarkable advances in multimodal GUI action generation grounded solely on screen snapshots~\citep{zhang2023appagentmultimodalagentssmartphone,wang2024mobileagentautonomousmultimodalmobile,hong2024cogagentvisuallanguagemodel,wu2024osatlasfoundationactionmodel,lin2024showuivisionlanguageactionmodelgui,qin2025uitarspioneeringautomatedgui,xu2025aguvisunifiedpurevision}.
Hence, \textit{a natural research question arises: how can we endow OpenClaw-style agents with the ability to efficiently interact with visual graphical environments?}

A straightforward approach is to integrate a standalone GUI agent into the OpenClaw framework. When deployed for long-horizon personal assistant tasks, this approach suffers from severe inefficiency and fragility due to four drawbacks: 
\textit{Firstly}, high-level user instructions frequently span multiple disjoint applications. Concise free-text summaries often fail to retain intermediate data values extracted from one app for subsequent cross-app operations. Conversely, rigidly enforcing explicit target application labels for vague tasks without accounting for device environments often leads to spurious app-allocation hallucinations.
\textit{Secondly}, GUI observations are inherently incomplete. Each individual data modality—screen captures, accessibility trees, foreground app IDs, historical action trajectories, and internal model reasoning logs—only reveals fragments of the full underlying device state. This necessitates the host agent to record historical trajectories and provide actionable guidance for the lightweight GUI Agent.
\textit{Thirdly}, successful and failed trajectories are typically discarded once a task terminates. When re-running analogous tasks, the agent is forced to re-launch target applications, redo redundant navigation steps, and re-learn known shortcuts and recurring failure patterns from scratch.
\textit{Finally}, most GUI workflows do not integrate faster non-visual shortcuts, such as web search tools, Android deep links, system intents, and reusable predefined action sequences. And such shortcuts cannot be safely repurposed as persistent long-term skills without validation against the real-time interface page.

This work presents \ourmodel, an agent framework augmented with structured knowledge and executable skills, built upon an OpenClaw-style host runtime and GUI-centric agent execution engine. The framework comprises two core functional components: an attribution policy-enhanced personalized memory system and a self-evolving skill library that supports rapid skill invocation and iterative optimization.
The design follows a simple principle: \emph{Knowing Deeply, Acting Perfectly}. Before acting, the agent retrieves task-relevant app candidates, tools, policies, and GUI hints from all agents' memories. During action, it treats GUI control as a partially observable decision process and records structured trajectory evidence or skills. After action, it converts useful traces into a skill library or generalized memory, so future runs can use operational knowledge rather than repeating exploratory behaviour. Specifically, \ourmodel treats GUI automation as a collaboration between a capable host agent and a lightweight GUI executor, not as a single monolithic GUI agent. The host owns the user-facing context---conversation, workspace memory, the user profile, external tools, and orchestration---and decides \emph{what} to do and which \emph{cli tool} could be used; the GUI executor owns screenshot perception, action normalization, device backends, skill validation, and trajectory recording, and decides \emph{how} to carry it out on screen. Framing the problem this way makes efficiency a first-class objective: \textbf{the system invokes the GUI subagent only when visual interaction is genuinely required, reuses validated operational skills and fast CLI tools whenever possible}.

On this foundation, \ourmodel contributes four connected mechanisms.
\begin{itemsize*}
    \item \textbf{Two-tier host--executor collaboration.} We design a structured, resumable GUI task interface. The host can assign limited GUI tasks to the executor and retrieve standardized outputs (fully completed, partially completed, or blocked), along with progress updates and resumption cues. Incomplete or stalled task runs act as reusable checkpoints: rather than sending one fuzzy command, the host either resumes unfinished execution or rearranges subsequent workflows. We also adopt an adaptive host involvement rule, where the host directly handles qualified information-gathering subtasks with tools instead of forwarding all subtasks to the GUI executor. This boosts task success rates and cuts computational overhead.
    \item \textbf{Memory-grounded routing and information transfer.} A built-in routing mechanism classifies user requests as single-app tasks or cross-app workflows spanning multiple applications. For cross-app sequences, every subtask explicitly defines its input and output data. A temporary shared data board transmits these structured values across subtasks, while persistent routing memory provides fixed candidate applications and auxiliary reference context.
    \item \textbf{Knowledge- and skill-augmented GUI execution.} \ourmodel distills trajectories into parameterized, state-validated skills, letting the executor commit to a reusable action prefix as a single decision. The same abstraction covers frequent click-then-type patterns and Android deeplink and intent shortcuts, which are taken only when their launch behavior and target page state satisfy the skill's contract.
    \item \textbf{Trajectory-derived memory and skill evolution.} Post-run reflection summarizes traces, distills success and failure lessons into retrievable experience memory, and refines the skill set as new evidence accumulates. This knowledge feeds back to the memory-grounded router before task decomposition and to the executor during execution, closing the loop from past runs to future decisions.
\end{itemsize*}

Extensive experimental results demonstrate that our model attains SOTA performance on the challenging MobileWorld benchmark, while enabling cross-platform deployment as well as memory and skill transfer.

\section{Related Work}
\label{sec:related_work}

\subsection{Personal Assistant Agent}

LLM-agent runtimes pair language models with action interfaces, memory, and control loops, as in ReAct~\citep{yao2023reactsynergizingreasoningacting}, the Agents framework~\citep{zhou2023agentsopensourceframeworkautonomous}, and OS-Copilot~\citep{wu2024oscopilotgeneralistcomputeragents}.Recently Two salient design philosophies recur when this pattern is packaged into local-first assistants. OpenClaw and Nanobot are configuration-centric: the host exposes message channels, external tools and Model Context Protocol (MCP) servers, conversation history and session state, long-lived workspace memory, and user-authored skills as first-class, declarable components, so that most behavior is specified around the model rather than left to it~\citep{openclaw2026,nanobot2026}. Hermes instead foregrounds adaptation, growing its competence over time by turning prior interactions into reusable routines that the assistant can later invoke~\citep{hermesagent2026}. Both directions show that a capable personal assistant benefits from treating control logic, memory, tools, and history as managed state around the model.

\subsection{GUI Agents}

GUI and computer-use agents study how language models act through visual interfaces rather than clean APIs. On the web, Mind2Web and WebArena expose long-horizon navigation across realistic sites through HTML and accessibility-tree observations~\citep{deng2023mind2webgeneralistagentweb,zhou2024webarenarealisticwebenvironment}, while VisualWebArena adds visually grounded decision making over rendered pages~\citep{koh2024visualwebarenaevaluatingmultimodalagents}, and SeeAct and WebVoyager explore screenshot- and page-structure-based control~\citep{zheng2024gpt4visiongeneralistwebagent,he2024webvoyagerbuildingendtoendweb}. Beyond the browser, Android in the Wild and AndroidWorld provide large-scale demonstrations and dynamic, programmatically checked tasks~\citep{rawles2023androidwildlargescaledataset,rawles2025androidworlddynamicbenchmarkingenvironment}, OSWorld and OmniACT cover broader desktop and web computer-use settings~\citep{xie2024osworldbenchmarkingmultimodalagents,kapoor2024omniactdatasetbenchmarkenabling}, and AppAgent and Mobile-Agent solve smartphone tasks with human-like actions and multimodal perception~\citep{zhang2023appagentmultimodalagentssmartphone,wang2024mobileagentautonomousmultimodalmobile}. Model-centric systems such as CogAgent, OS-Atlas, ShowUI, UI-TARS, and Aguvis further improve perception, grounding, and visual action generation~\citep{hong2024cogagentvisuallanguagemodel,wu2024osatlasfoundationactionmodel,lin2024showuivisionlanguageactionmodelgui,qin2025uitarspioneeringautomatedgui,xu2025aguvisunifiedpurevision}.These works have substantially improved GUI task execution, their primary focus is often on task completion within benchmarked interaction environments. 

\subsection{Memory and Skill Reuse}

Agent memory and skill reuse let agents improve without weight updates \citep{vu2018sentence} in both language-only and embodied settings. Reflexion stores verbal feedback to guide later attempts~\citep{shinn2023reflexionlanguageagentsverbal}, Generative Agents maintain a reflective memory stream for coherent long-term behavior~\citep{park2023generativeagentsinteractivesimulacra}, Voyager builds an executable skill library for open-ended embodied learning~\citep{wang2023voyageropenendedembodiedagent}, and KnowAgent uses an action knowledge base to curb planning hallucination~\citep{zhu2025knowagentknowledgeaugmentedplanningllmbased}. ReasoningBank distills reusable success and failure rationales into retrievable memory for self-evolving agents~\citep{ouyang2026reasoningbankscalingagentselfevolving}, GUI-specific work such as LearnAct and CUA-Skill highlights the value of demonstrations, retrieved knowledge, and engineered skills for computer-use agents~\citep{liu2025learnactfewshotmobilegui,chen2026cuaskilldevelopskillscomputer}, and a recent survey frames agent memory as a write--manage--read loop coupled with perception and action~\citep{du2026memoryautonomousllmagentsmechanisms}. Across these lines, textual memory and executable skills address related but distinct reuse problems: memory preserves guidance, rationales, and context, whereas skills compress procedures into reusable behavior.

\section{Preliminaries}
\label{sec:preliminaries}

\subsection{GUI Automation as a POMDP}

GUI automation is naturally modeled as a partially observable Markov decision process (POMDP)~\citep{kaelbling1998planning}:
\begin{equation}
\mathcal{M}=(\mathcal{S}, \mathcal{A}, \mathcal{O}, T, \Omega, R, \gamma),
\end{equation}
where $\mathcal{S}$ is the hidden device state, $\mathcal{A}$ the action space, $\mathcal{O}$ the observation space, $T$ and $\Omega$ the transition and observation functions, $R$ the task reward, and $\gamma$ the discount factor. The defining property of GUI control is partial observability: the agent never observes $\mathcal{S}$ directly but only an observation $o_t$---typically a screenshot with optional screen metadata, the foreground app, and a bounded action history---that exposes a projection of hidden state such as navigation stacks, login and permission status, asynchronous loading, and off-screen form values. A GUI agent must therefore act and recover from errors under this uncertainty rather than assume a fully known interface.

\subsection{Host-Centric Multi-Agent Systems}

Multi-agent systems distribute problem solving across agents with distinct roles, capabilities, and local information, coordinating through communication and shared environments rather than a single monolithic policy~\citep{wooldridge2009introduction}. Recent LLM agent frameworks adopt this view, pairing a central orchestrator that maintains conversation state, memory, and tool access with specialized environment- or tool-facing agents to which it delegates subtasks~\citep{zhou2023agentsopensourceframeworkautonomous}. For GUI automation, this host-orchestration pattern motivates separating long-horizon task management from low-level interaction with a partially observed device; Section~\ref{sec:methodology} instantiates this as a concrete execution stack for mobile GUI tasks.

\section{KnowAct-GUIClaw}
\label{sec:methodology}

\subsection{Overview: A Know--Route--Act--Reflect Stack}

\begin{figure}[t]
    \centering
    \includegraphics[width=\linewidth]{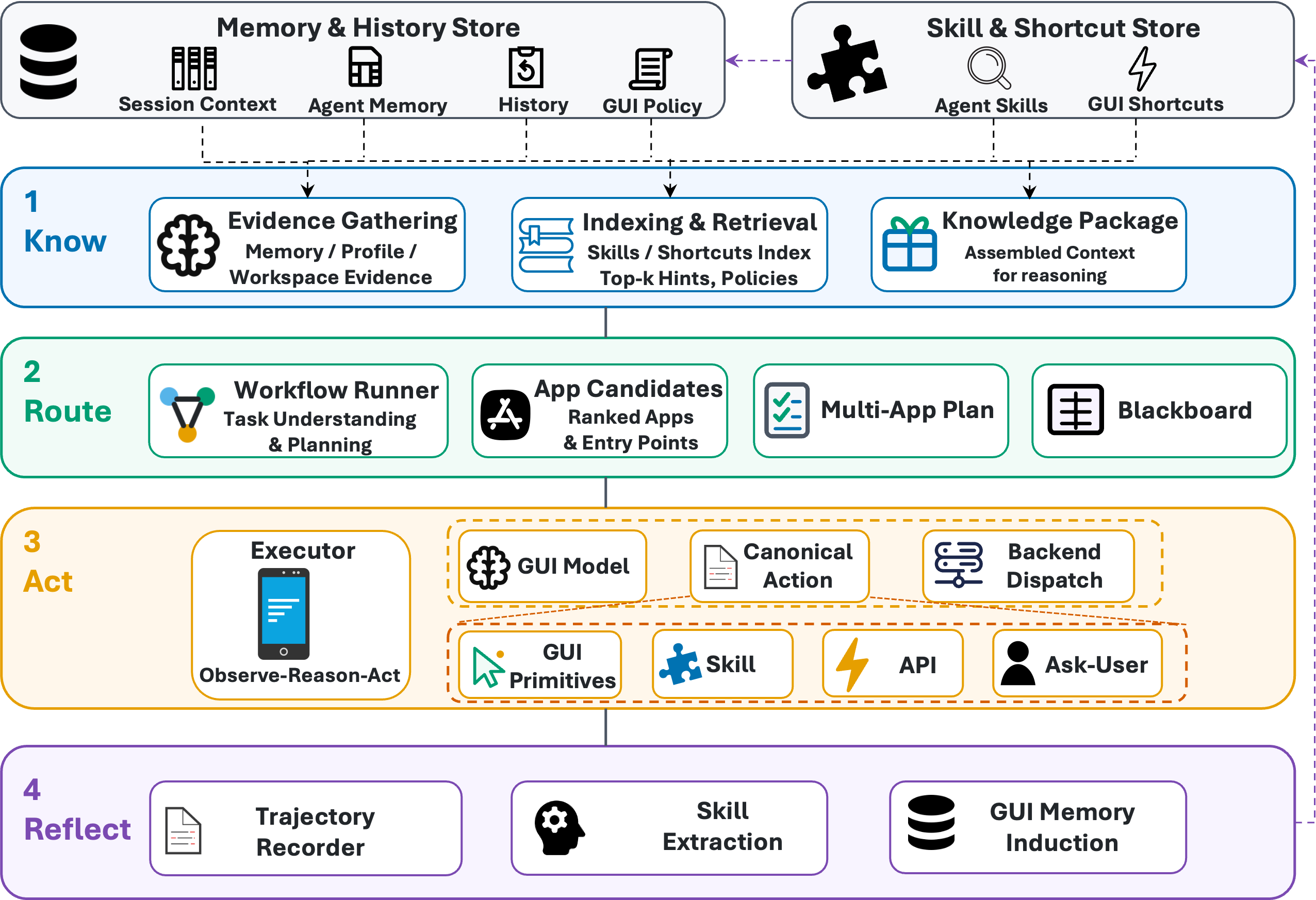}
    \caption{\textbf{Overview of the \ourmodel execution loop}. Two persistent stores---a memory and history store and a skill and shortcut store---supply advisory context to every stage. \emph{Know} gathers evidence and assembles a reasoning context; \emph{Route} ranks app candidates and turns the request into either a single GUI task or an ordered multi-app workflow whose subtasks exchange typed values through a blackboard; \emph{Act} runs GUIClaw's observe--reason--act loop over the hybrid action space of GUI primitives, skills, deeplink/intent shortcuts, and intervention actions; and \emph{Reflect} distills each trajectory into updated skills and experience memory that feed back into the stores.}
    \label{fig:method_overview}
\end{figure}

Recent mobile-agent systems and benchmarks point to three requirements: hybrid GUI--shortcut action spaces improve efficiency when shortcuts are available~\citep{zhao2026masbenchunifiedbenchmarkshortcutaugmented}; perception, memory, and action work as one execution stack~\citep{ren2026xomniclawtechnicalreportunified}; and a GUI-agent framework should learn from the execution trajectories~\citep{tang2026clawgui}. \ourmodel organizes new long-horizon task execution as a four-stage loop: \emph{Know}, \emph{Route}, \emph{Act}, and \emph{Reflect}. Figure~\ref{fig:method_overview} summarizes this loop and the persistent stores that feed every stage.

GUI execution should serve as a subagent within the personal user assistant framework, specifically to address scenarios where no standardized API endpoints are available and graphical interfaces are dynamically rendered in real time. Accordingly, the GUI task interface acts as the formal boundary between the host agent and the UI subagent. Each GUI task returns a structured output consisting of three components: its completion status, a concise execution summary, and the final screen state upon task termination. The host agent retains responsibility for user-facing task orchestration and long-horizon context maintenance, while the GUI subagent operates as a self-contained engine dedicated to low-level device control.

\subsection{Know: Context Gathering and Host Control}
\label{sec:method_host_policy}

\begin{figure}[t]
    \centering
    \includegraphics[width=\linewidth]{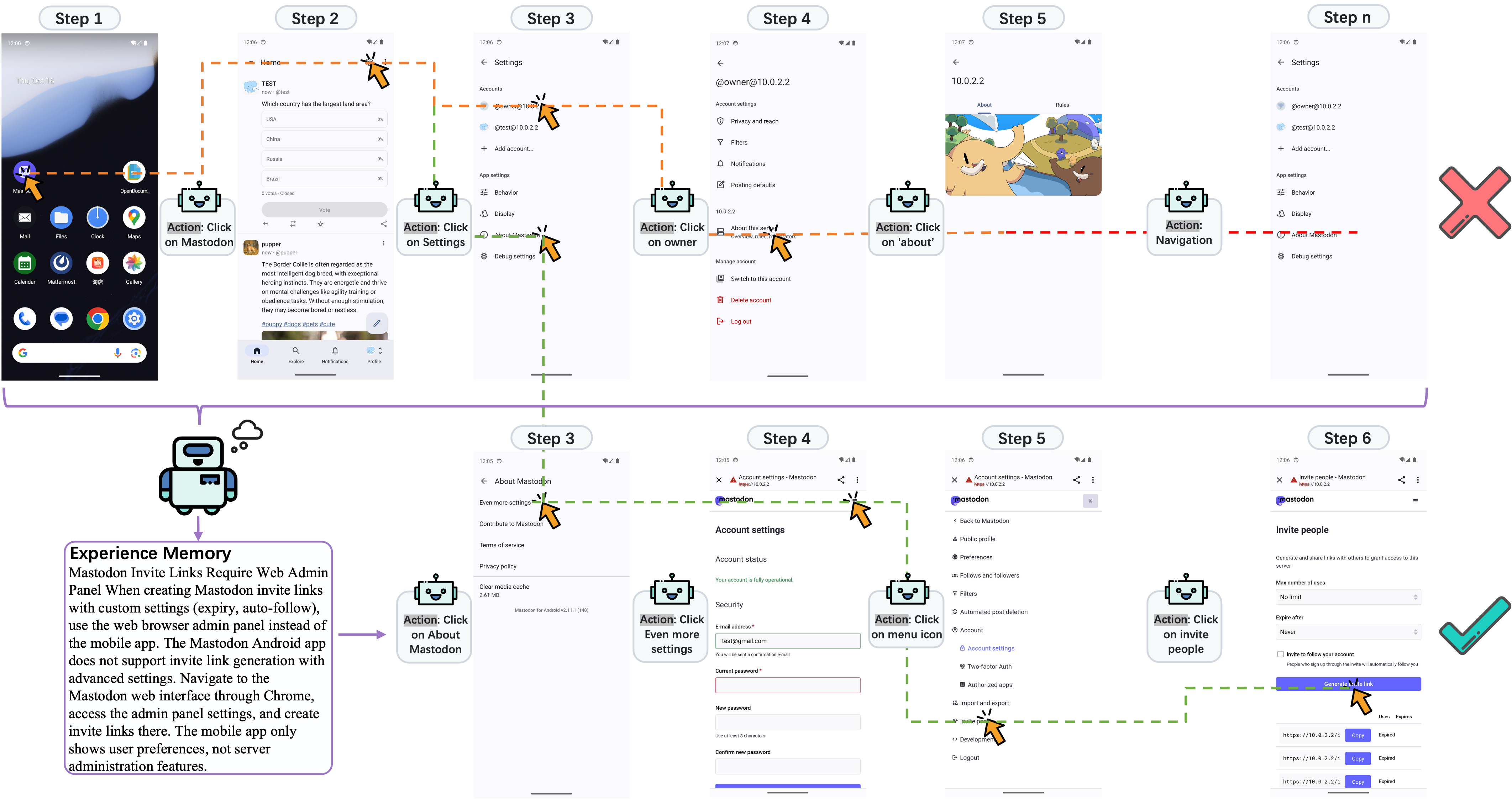}
    \caption{\textbf{Experience memory improves a GUI task by changing the task context before low-level control begins}. Without the retrieved memory (Top), GUIClawinvites continues through Mastodon's mobile settings and reaches a nonproductive path for invite-link creation. With the retrieved memory (bottom), the Know stage supplies an advisory lesson that invite links with advanced settings that require the web administration panel; GUIClaw then opens the web interface, navigates to account settings, and reaches the invite-people page. The example shows that experience memory guides app choice, decomposition, and recovery while live screen observations still ground each action.}
    \label{fig:gui-memory-case}
\end{figure}

\noindent\textbf{Active retrieval and policy injection.} Before any GUI action, the Know stage gathers most of its context actively: prior GUI memories and candidate skills are retrieved by semantic similarity and kept advisory, never overriding the current instruction, while policy memory is injected directly rather than ranked. Figure~\ref{fig:gui-memory-case} illustrates this advisory role on a Mastodon task: the retrieved lesson redirects the task from unsupported mobile-app settings to the web administration panel without replaying an old trajectory.

\noindent\textbf{Host-held context and active recall.} The host agent keeps the running session context and forwards only what a GUI task needs when it issues that task, and it recalls its own stored memories, such as session history, agent memory, and the user profile, only when it judges them relevant. When the inputting instruction is fuzzy, it draws on these memories to propose defaults as explicit, overridable assumptions. \textit{This capability is critical for an agentic assistant to achieve continuous capability improvement when deployed across mobile and desktop platforms.}

\noindent\textbf{Host-centric Control.} The host agent delegates selectively. It answers a subtask itself when conversation context, recalled memory, or non-GUI tools suffice and no device-local GUI state must be observed or changed, and it issues a GUI task only when the subtask needs a live app session, visual grounding, cross-app manipulation, text entry, or on-device verification. Resolving eligible subtasks directly reduces hand-offs and token-heavy GUI traces.

\subsection{Route: Task Decomposition and Information Contracts}
\label{sec:method_router}

\noindent\textbf{Task decomposition.} The routing policy emits either a single GUI task or an ordered multi-app workflow. In the multi-app case, each subtask is a goal-level tuple $(g_i,h_i,I_i,O_i)$: $g_i$ is the app-scoped goal, $h_i$ optionally narrows the app, $I_i$ names the inputs it requires, and $O_i$ names the values it should return. The router does not predict screen sequences. Because each subtask has its corresponding app, the relevant memory and skills are retrieved afresh for that subtask rather than inherited from the top-level route.

\begin{figure}[t]
    \centering
    \includegraphics[width=\linewidth]{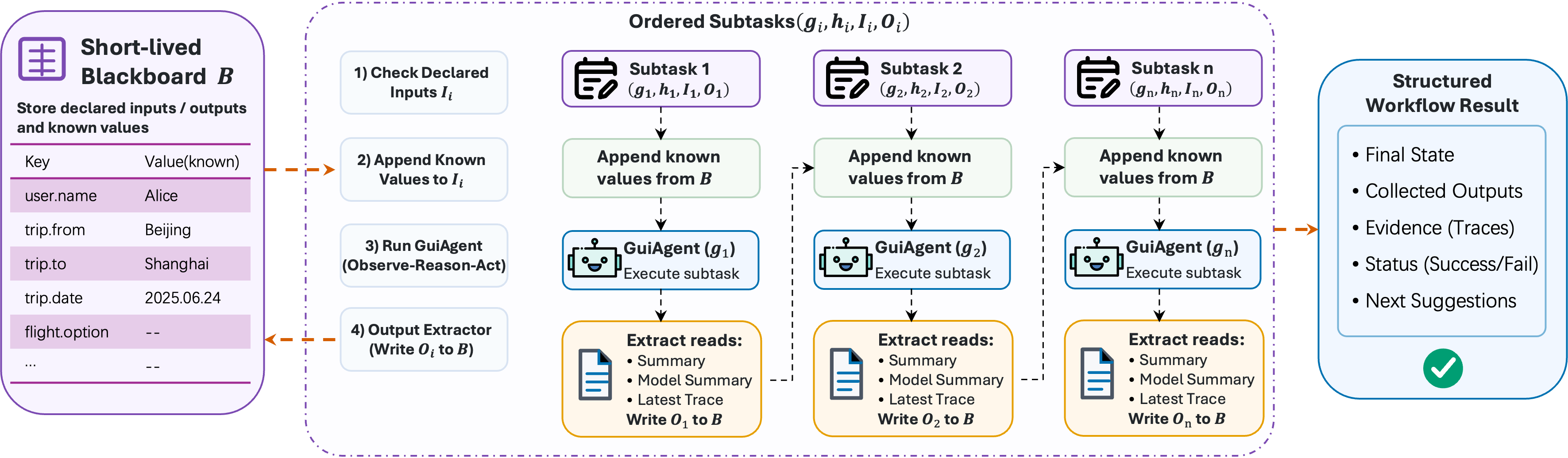}
    \caption{\textbf{Blackboard-mediated execution in the Route stage}. The short-lived blackboard $B$ stores typed inputs and outputs known so far. Each subtask $(g_i,h_i,I_i,O_i)$ checks its declared inputs $I_i$, reads their values from $B$, runs GUIClaw's observe--reason--act loop to produce a trajectory $\tau_i$, and writes only its declared outputs $O_i$ back to $B$~\eqref{eq:blackboard}. A missing required input or output makes the workflow fail closed, so later subtasks consume observed typed values rather than free-form summaries.}
    \label{fig:blackboard}
\end{figure}

\noindent\textbf{Information transfer.} The blackboard makes cross-app data flow explicit. Let $G$ denote GUIClaw, which executes a subtask over the POMDP, let $E$ denote the evidence-to-value mapping, and let $B_i$ denote the blackboard after subtask $i$. A subtask sees only the declared inputs already on $B_{i-1}$, and only its declared outputs are written back from the trajectory evidence:
\begin{equation}
\label{eq:blackboard}
G(g_i,h_i,B_{i-1}[I_i]) \rightarrow \tau_i,\qquad
E(\tau_i,O_i) \rightarrow B_i[O_i].
\end{equation}
If a required input or a declared output is missing, the workflow fails closed rather than running a subtask on incomplete state or letting a later one infer or fabricate the value. This rule matters for cross-app transfer, comparison, and fact-collection tasks. Figure~\ref{fig:blackboard} traces the typed tuple contract, while Figure~\ref{fig:case1-trajectory} (Section~\ref{sec:experiments}) shows the same mechanism in a real cross-app trajectory.

\subsection{Act: Hybrid GUI--Fast Path Execution}
\label{sec:method_skills}

\noindent\textbf{Hybrid action space.} GUIClaw executes each routed subtask over a hybrid action space:
\begin{equation}
\mathcal{A}=
\mathcal{A}_{\mathrm{gui}}
\cup \mathcal{A}_{\mathrm{skill}}
\cup \mathcal{A}_{\mathrm{shortcut}}
\cup \mathcal{A}_{\mathrm{ask}} .
\end{equation}
$\mathcal{A}_{\mathrm{gui}}$ contains human-like primitives such as tap, swipe, scroll, text input, navigation, app open/close, and wait; $\mathcal{A}_{\mathrm{skill}}$ contains skills distilled from reusable GUI behavior. $\mathcal{A}_{\mathrm{shortcut}}$ contains Android deeplinks and intents~\citep{androiddeeplinks2026,androidintents2026} which bypass lengthy navigation workflows when verified as valid on the target device. Predefined action sequences with historically calibrated interaction parameters can also be stored in this set to eliminate the visual grounding overhead of the GUI subagent, though such rigid shortcut schemes suffer from limited generalization capability.  $\mathcal{A}_{\mathrm{ask}}$ contains intervention actions for cases that require user input or authorization. These types trade efficiency against reliability, so the executor validates skills and shortcuts against the current state before using them. Appendix~\ref{app:action-space} gives the unified action space, by platform, that realizes $\mathcal{A}_{\mathrm{gui}}\cup\mathcal{A}_{\mathrm{ask}}$.

\noindent\textbf{Observe--Reason--Act loop.} Within this action space, GUIClaw runs a grounded control loop:
\begin{equation}
p_t \rightarrow o_t \rightarrow a_t \rightarrow e_t \rightarrow o_{t+1},
\end{equation}
where $p_t$ is the prompt, $o_t$ is the current observation, $a_t$ is the action, and $e_t$ its result on the device. The prompt carries the subtask, task's blackboard, policy memory, interface hints, and available skills, and a normalization step maps each model's output into a common action format. Appendix~\ref{app:skill-induction} gives the prompt contracts used to expose, induce, and validate these skills.

\noindent\textbf{Skills.} Skills turn repeated interaction into reusable procedures. Each stored skill contains an identifier, app and platform scope, description, parameters, reliability counters, and an ordered sequence of steps. Stable fields such as app package names, coordinates, components, and intent payloads are fixed in advance, while task-dependent values are represented as placeholders and grounded at run time. Before each subtask, the GUI subagent retrieves app-scoped candidates by lexical and embedding signals, then uses a lightweight applicability prompt to select one useful prefix or reject all candidates. Retrieved skills (Top-5) are advisory: \ourmodel may apply one or fall back to ordinary actions. Before each step, it checks the expected state through a deterministic state contract when available and through a visual valid-state check otherwise; on a mismatch, it either runs a bounded recovery subgoal, skips an optional obstacle step, or returns control to ordinary GUI execution. Appendix~\ref{app:skills} gives concrete skill records, including a multi-step skill with per-step state validation.

\noindent\textbf{Deeplink and intent shortcuts.} $\mathcal{A}_{\mathrm{shortcut}}$ contains page-validated Android deeplinks and intents that bypass long GUI navigation. During execution, GUIClaw treats them as one-step skills with \texttt{open\_deeplink} or \texttt{open\_intent}, and uses them only after their target page, required parameters, and app state have been validated. This prevents broad Android constants such as \texttt{SEARCH}, \texttt{SEND}, or \texttt{GET\_CONTENT} from being executed merely because a manifest exposes them. Appendix~\ref{app:skills} shows representative skill and shortcut entries.

\subsection{Reflect: Trajectory Distillation and Skill Evolution}
\label{sec:method_memory}

\noindent\textbf{Post-run summarization.} After each GUI task, reflection condenses the trajectory into a short note of where execution ended and what remains, so a partial or blocked task becomes a checkpoint the host can resume without redoing finished work. The post-run processor summarizes the trace and, when enabled, evaluates the outcome before learning from it. Runs with no useful signal, e.g., empty, cancelled, timed out before progress, or completed entirely by an already reused skill, are not written to the long-term stores.

\noindent\textbf{Skill extraction.} For trajectories selected for skill learning, reflection extracts a reusable procedure rather than storing raw actions for replay. It rewrites GUI events into structured evidence: the task, platform, observed apps, action sequence, action parameters, screenshots, target-control hints, and inferred state contracts. A vision LLM receives this evidence under a restricted prompt that permits only declarative action sequences. The candidate skill is normalized and checked for supported actions, declared parameters, executable fixed fields, reusable app scope, and valid-state coverage. Accepted steps inherit the state contracts inferred from the trajectory, so later execution validates screen state instead of replaying an ungrounded script. Offline extraction uses the same agent after filtering short or abnormal traces, keeping bounded successful prefixes, clustering structurally identical skills, and recording success counts for retrieval priority.

\noindent\textbf{Skill evolution.} \ourmodel separates repair from new extraction. When a trajectory shows a reused skill failing, reflection records the failed step, screen evidence, error, original skill, and prior feedback, then asks an evolution prompt to update that same skill in place: it may narrow the description, add guarded optional obstacle handling, or refresh stale targets and state contracts, but it may not replace the skill with an unrelated workflow or add destructive terminal actions. Repair therefore takes precedence over new extraction; only when no failed reused skill is present does the system mine a new skill from reusable behavior.

\noindent\textbf{Shortcut validation.} Shortcut candidates are mined from app manifests as discovery evidence, not trusted actions. A validation run launches candidate variants, records the foreground app, ADB output, UI tree, and screenshot, and asks a verifier to return usability, page status, payload preservation, parameters, and a natural-language capability description. Only page-validated records are promoted into one-step skills; merely launchable records remain candidates unless explicitly allowed by the validation configuration. Appendix~\ref{app:skill-induction} gives the validation prompt.

\noindent\textbf{Experience memory.} Experience memory stores textual policies derived from running trajectories. Inspired by ReasoningBank \citep{ouyang2026reasoningbankscalingagentselfevolving}, \ourmodel formats each trace into a concise action-and-UI summary and uses separate success and failure prompts to induce at most a few actionable memory items. The memory inducer skips short or abnormal traces, resolves outcome and app at the per-trace level, caps the number of retained items per task, and drops near-duplicates within the same app. These lessons are distinct from executable skills and feed later tasks at both levels: routing draws on them to choose apps and decomposition patterns, while the GUI agent uses them as hints about layout, shortcut reliability, and recovery. Appendix~\ref{app:skill-induction} gives the corresponding induction prompts and filtering rules.

\section{Experiments}
\label{sec:experiments}

\subsection{Benchmarks and Metrics}

We evaluate \ourmodel on two mobile GUI benchmarks that include diverse real-user tasks. \emph{MobileWorld}~\citep{kong2025mobileworldbenchmarkingautonomousmobile} is our primary benchmark: it emphasizes long-horizon mobile tasks, many of which span multiple applications. From its 201 tasks over 20 applications, we use the 117-task GUI-Only subset, scored by the benchmark's native deterministic evaluators. \emph{AndroidDaily}~\citep{yan2025stepguitechnicalreport} is a complementary end-to-end benchmark that groups tasks by type, complexity, and ambiguity; many of its tasks require returning an explicit answer rather than only manipulating the UI. Lacking a native evaluator, we score its GUI-only tasks with \texttt{qwen3.5-flash} as an LLM judge and its answer-returning tasks with two human experts, with correct tasks scoring $1.0$ and partial tasks $0.5$.

\noindent\textbf{Setup.} Unless noted, \ourmodel pairs a Qwen3.5-397B-A17B host with a Qwen3.5-35B-A3B GUI executor. In the Kimi-K2.6 configurations of Table~\ref{tab:mobileworld-leaderboard}, Kimi-K2.6 serves as both the host and GUI executor. The Kimi-to-Qwen transfer configuration retains the default Qwen host--executor pairing and instead uses a joint set of experience memory and skills distilled from trajectories generated by the Kimi-K2.6 host--executor configuration. All MobileWorld configurations use the same 117-task GUI-Only subset, native deterministic evaluators, and 50-step cap. Following the public leaderboard, our primary MobileWorld metric is the single-run success rate (SR), i.e.\ pass@1; we report pass@3 only as a repeated-attempt upper bound, in a separate table (Table~\ref{tab:pass3}). For AndroidDaily we report a \emph{resolved} setting over the 194 available tasks and an \emph{all} setting that scores the 41 unavailable entries as $0$ across all 235 tasks. The AndroidDaily evaluation uses iOS devices.

\noindent\textbf{Efficiency metrics.} We report SR together with execution cost: GUI steps, the number of GUI task invocations, executed GUI-trace tokens, and host tokens. The \emph{total} column measures the executed GUI traces and is comparable across systems; \emph{host total} measures the additional generation that the host spends to route, coordinate, call external tools such as web search, and resolve eligible subtasks.

\subsection{Main Results of MobileWorld}

\begin{table}[t]
    \caption{MobileWorld GUI-Only success rate (pass@1). External rows are grouped by the public MobileWorld leaderboard categories~\citep{kong2025mobileworldbenchmarkingautonomousmobile}. All \ourmodel rows are our own runs; the Kimi-to-Qwen transfer row equips the 35B executor with experience memory and skills distilled from Kimi-K2.6 trajectories. Rows matching configurations B, C, and F are analyzed in Table~\ref{tab:efficiency}, and Table~\ref{tab:pass3} reports the repeated-attempt upper bound.}
    \label{tab:mobileworld-leaderboard}
    \centering
    \small
    \begin{tabular}{@{}p{0.80\linewidth}r@{}}
        \toprule
        Model & SR (pass@1, \%) \\
        \midrule
        \multicolumn{2}{@{}l}{\textit{General models}} \\
        Seed-2.0-Pro~\citep{seed2026seed20modelcardintelligence} & 63.2 \\
        GPT-5.5~\citep{openai2026gpt55} & 62.4 \\
        Gemini-3.1-Pro-Preview~\citep{google2026geminipro} & 58.1 \\
        Claude-Opus-4.7~\citep{anthropic2026claudeopus47} & 56.4 \\
        Kimi-K2.6~\citep{moonshot2026kimik26} & 55.6 \\
        Kimi-K2.5~\citep{moonshot2026kimik25} & 49.6 \\
        Claude-Sonnet-4.5~\citep{anthropic2025claudesonnet45} & 47.8 \\
        Qwen3.5-397B-A17B~\citep{qwen2026qwen35} & 42.7 \\
        Qwen3.5-122B-A10B~\citep{qwen2026qwen35} & 35.0 \\
        Qwen3.5-35B-A3B~\citep{qwen2026qwen35} & 24.8 \\
        Qwen3-VL-235B-A22B~\citep{bai2025qwen3vltechnicalreport} & 12.8 \\
        \addlinespace
        \multicolumn{2}{@{}l}{\textit{Agentic systems}} \\
        Gemini-3-Pro + UI-Ins-7B~\citep{google2026geminipro,chen2025uiinsenhancinggui} & 55.6 \\
        GPT-5 + UI-Ins-7B~\citep{singh2026openaigpt5card,chen2025uiinsenhancinggui} & 54.0 \\
        Claude-4.5-Sonnet + UI-Ins-7B~\citep{anthropic2025claudesonnet45,chen2025uiinsenhancinggui} & 47.8 \\
        \addlinespace
        \multicolumn{2}{@{}l}{\textit{Specialized GUI models}} \\
        GUI-Owl-1.5-32B~\citep{xu2026mobileagentv35multiplatformfundamentalgui} & 43.9 \\
        ForgeOwl-8B~\citep{liu2026mobileforgeannotationfreeadaptationmobile} & 41.0 \\
        MAI-UI-235B-A22B~\citep{zhou2025maiuitechnicalreportrealworld} & 39.7 \\
        GUI-Owl-1.5-8B~\citep{xu2026mobileagentv35multiplatformfundamentalgui} & 38.2 \\
        MAI-UI-32B~\citep{zhou2025maiuitechnicalreportrealworld} & 36.2 \\
        Doubao-1.5-UI-TARS~\citep{qin2025uitarspioneeringautomatedgui} & 26.3 \\
        UI-Venus-72B~\citep{gu2025uivenustechnicalreportbuilding} & 16.4 \\
        GUI-Owl-7B~\citep{ye2025mobileagentv3fundamentalagentsgui} & 7.7 \\
        \midrule
        \multicolumn{2}{@{}l}{\textit{Ours (Open-source Qwen3.5)}} \\
        \rowcolor[HTML]{E3F2FD}
        \textbf{35B +\,host \& memory} & \textbf{34.5} \\
        \rowcolor[HTML]{E3F2FD}
        \textbf{35B +\,host, memory \& skills} & \textbf{37.9} \\
        \rowcolor[HTML]{E3F2FD}
        \textbf{35B +\,host, Kimi-derived memory \& skills} & \textbf{41.0} \\
        \rowcolor[HTML]{E3F2FD}
        \textbf{397B host acts directly} & \textbf{46.2} \\
        \addlinespace
        \multicolumn{2}{@{}l}{\textit{Ours (Open-source Kimi-K2.6)}} \\
        \rowcolor[HTML]{E0F7FA}
        \textbf{+\,host \& memory} & \textbf{61.5} \\
        \rowcolor[HTML]{E0F7FA}
        \textbf{+\,host, memory \& skills} & \textbf{64.1} \\
        \bottomrule
    \end{tabular}
\end{table}

Table~\ref{tab:mobileworld-leaderboard} compares \ourmodel with the public MobileWorld leaderboard. The Qwen3.5-35B-A3B GUI executor reaches 24.8 SR on its own; running it inside \ourmodel with the host and experience memory raises pass@1 to 34.5, and enabling skills on top raises it to 37.9. Equipping the same 35B executor with experience memory and skills distilled from Kimi-K2.6 trajectories further raises SR to 41.0, a 16.2-point gain over the base executor and 3.1 points above the standard 35B host--memory--skills configuration. With Kimi-K2.6 as the base model, \ourmodel reaches 61.5 with the host and experience memory and 64.1 after enabling skills, the highest SR in the table. Letting the 397B Qwen host act directly reaches 46.2; using only open Qwen3.5 models, this configuration remains above the public Qwen3.5-397B-A17B plain-GUI score (42.7) and every specialized end-to-end GUI model listed. We isolate the host--memory and skills increments next.

\subsection{Ablation and Efficiency Analysis}

\begin{table}[t]
    \caption{MobileWorld GUI-Only ablation and efficiency (our runs). Rows A--C use the Qwen3.5-35B-A3B GUI executor and rows D--F the Qwen3.5-397B-A17B; \textbf{+\,host \& mem} adds the \ourmodel host, router, and experience memory, and \textbf{+\,skills} enables skills on top of that setting. In E--F the 397B host resolves eligible subtasks directly rather than delegating each to the 35B executor (Section~\ref{sec:method_host_policy}). Row D is our plain-GUI 397B run, distinct from its 42.7 public leaderboard score. \emph{Total} is GUI-trace tokens per task; \emph{host total} is the host's own generation, including external tool calls, and is the extra token cost \ourmodel adds on top. Highlighted rows mark the host- and skill-augmented configurations.}
    \label{tab:efficiency}
    \centering
    \small
    \resizebox{0.78\linewidth}{!}{
    \begin{tabular}{llccccc}
        \toprule
        Config & Setting & SR (\%) & GUI steps & GUI tasks & Total & Host Total \\
        \midrule
        \multicolumn{7}{l}{\textit{GUI executor: Qwen3.5-35B-A3B}} \\
        A & exec.\ only & 24.8 & 26.7 & 1.0 & 281{,}266 & -- \\
        \rowcolor[HTML]{E3F2FD}
        B & +\,host \& mem & 34.5 & 26.8 & 2.3 & 279{,}211 & 65{,}224 \\
        \rowcolor[HTML]{E3F2FD}
        C & +\,skills & 37.9 & 25.1 & 2.5 & 278{,}289 & 63{,}792 \\
        \midrule
        \multicolumn{7}{l}{\textit{GUI executor: Qwen3.5-397B-A17B}} \\
        D & exec.\ only & 40.7 & 26.1 & 1.0 & 254{,}459 & -- \\
        \rowcolor[HTML]{E3F2FD}
        E & +\,host \& mem & 43.3 & 26.8 & 1.4 & 273{,}352 & 10{,}096 \\
        \rowcolor[HTML]{E3F2FD}
        F & +\,skills & \textbf{46.2} & \textbf{23.7} & 1.7 & \textbf{260{,}516} & 10{,}982 \\
        \bottomrule
    \end{tabular}}
\end{table}

Table~\ref{tab:efficiency} reports a system-level ablation along two executor scales.

\noindent\textbf{Host and memory raise accuracy at a modest comparable cost.} Adding the host, router, and experience memory raises SR at both executor scales (A$\to$B, $24.8\to34.5$; D$\to$E, $40.7\to43.3$). Total tokens stay flat for the 35B executor ($281$k vs.\ $279$k) but rise about $7\%$ for the 397B executor ($254$k to $273$k), where the host splits a task into more GUI subtrajectories. Because total tokens are dominated by visual observation, the accuracy gain reflects better task organization rather than a larger GUI-executor budget; the host total is $65{,}224$ tokens per task when the host must coordinate the weaker 35B executor across $2.3$ GUI task invocations (about $23\%$ of total), but only $10{,}096$ with the 397B host at $1.4$ invocations (under $4\%$).

\noindent\textbf{Skills cut steps and total tokens (B$\to$C, E$\to$F).} Enabling skills on top of the host--memory setting lowers GUI steps and total tokens---modestly for the 35B executor, and more clearly for the 397B host, where steps fall from $26.8$ to $23.7$ and total tokens by about $5\%$---while SR rises rather than falls. The saving comes from shorter trajectories and fewer screenshot-heavy observations. Table~\ref{tab:skill-reuse} isolates this effect on the tasks that actually invoke a skill.

\noindent\textbf{Experience memory and skills transfer across base models.} Table~\ref{tab:mobileworld-leaderboard} includes a cross-model transfer test in which the Qwen3.5-35B-A3B executor uses experience memory and executable skills distilled from Kimi-K2.6 trajectories. This transferred configuration reaches 41.0 SR, compared with 37.9 for the standard 35B host--memory--skills configuration and 24.8 for the base executor. The result shows that both textual experience rationales and state-validated executable behaviors can carry task knowledge across model families rather than remaining tied to the model that generated the source trajectories. Together with the Kimi-K2.6 results (61.5 with host and memory; 64.1 with skills), this establishes transferability in the tested Kimi-to-Qwen direction without assuming universal model-independent transfer.

\noindent\textbf{A capable host should not route everything through the GUI (F).} Letting the 397B host resolve eligible information-query subtasks directly, rather than delegating each to the 35B executor, yields the best SR ($46.2$) while lowering both cost components relative to the delegated configuration C: the host answers eligible subtasks itself instead of spawning a GUI task for each ($1.7$ GUI task invocations versus $2.5$ for C), so total tokens fall to $260{,}516$ (from $278{,}289$) and host total to $10{,}982$ tokens per task (from $63{,}792$). F thus delivers the best accuracy at the lowest GUI-trace execution cost among the Qwen ablations, while keeping host overhead minimal. For external context, Qwen3.7-Plus reaches $43.6$ SR with $30.0$ steps, while the public Kimi-K2.6 baseline reaches $55.6$; our Kimi host--memory and skill-augmented configurations reach $61.5$ and $64.1$, respectively (Table~\ref{tab:mobileworld-leaderboard}).

\begin{table}[t]
    \caption{Skill-reuse effect on the MobileWorld tasks that invoke at least one skill ($83$ with the 35B executor, $87$ with the 397B host), each compared against the same tasks run without skills. SR and pass@3 deltas are in percentage points.}
    \label{tab:skill-reuse}
    \centering
    \small
    \resizebox{0.85\linewidth}{!}{
    \begin{tabular}{lcccccc}
        \toprule
        & \multicolumn{3}{c}{Qwen3.5-35B-A3B executor} & \multicolumn{3}{c}{Qwen3.5-397B-A17B executor} \\
        \cmidrule(lr){2-4} \cmidrule(lr){5-7}
        Metric (skill-using tasks) & +\,skills & no skills & $\Delta$ & +\,skills & no skills & $\Delta$ \\
        \midrule
        GUI steps / task            & 25.7 & 29.0 & $-3.3$ & 22.3 & 25.6 & $-3.3$ \\
        total tokens / task         & 284{,}279 & 303{,}014 & $-6.2\%$ & 242{,}930 & 258{,}084 & $-5.9\%$ \\
        single-run SR (\%)          & 40.6 & 35.7 & $+4.9$ & 50.2 & 48.3 & $+1.9$ \\
        pass@3 (\%)                 & 54.2 & 53.0 & $+1.2$ & 63.2 & 63.2 & $+0.0$ \\
        \bottomrule
    \end{tabular}}
\end{table}

\noindent\textbf{Skill reuse, measured where it applies.} Restricting to the tasks that invoke a skill concentrates the same effect (Table~\ref{tab:skill-reuse}): reuse removes about three GUI steps per task ($-3.3$), cuts total and prompt tokens by roughly $6\%$, and improves SR ($+4.8$ and $+1.9$ points).

Section~\ref{sec:case-study} complements these aggregate comparisons with qualitative evidence for routing and information transfer, experience-memory correction, navigation compression through skills and shortcuts, and host recovery and direct participation.

\FloatBarrier

\subsection{AndroidDaily Results}

\begin{table}[!ht]
    \caption{AndroidDaily (end-to-end) performance by task type, complexity, and ambiguity. Our AndroidDaily evaluation is conducted on iOS devices. For \ourmodel, \emph{resolved} excludes the 41 unavailable entries and rates the remaining 194 tasks, while \emph{all} scores those entries as $0$ over all 235 tasks.}
    \label{tab:androiddaily-completion}
    \centering
    \small
    \resizebox{\linewidth}{!}{
    \begin{tabular}{lcccccccccc}
        \toprule
        & \multicolumn{3}{c}{Task Type} & \multicolumn{3}{c}{Complexity} & \multicolumn{3}{c}{Ambiguity} & \\
        \cmidrule(lr){2-4} \cmidrule(lr){5-7} \cmidrule(lr){8-10}
        Model / Setting & Filter & Query & Analyze & Atomic & Comp. & Cond. & Low & Mid & High & Total \\
        \midrule
        UI-TARS-1.5 & 57.64 & 65.97 & 36.71 & 61.41 & 13.64 & 60.38 & 57.05 & 54.90 & 57.89 & 56.64 \\
        Step-GUI-4B & 44.77 & 64.29 & 33.72 & 54.03 & 19.61 & 42.86 & 51.21 & 38.32 & 59.52 & 49.06 \\
        Step-GUI-8B & 52.50 & 63.82 & 32.95 & 59.09 & 14.00 & 42.86 & 54.08 & 44.55 & 61.54 & 52.50 \\
        \midrule
        \rowcolor[HTML]{E3F2FD}
        Ours (Resolved) & 80.56 & 76.88 & 77.08 & 81.25 & 62.50 & 75.00 & 78.89 & 77.50 & 78.95 & \textbf{78.61} \\
        \rowcolor[HTML]{E3F2FD}
        Ours (All) & 68.40 & 66.13 & 51.39 & 70.98 & 41.67 & 53.23 & 64.94 & 65.96 & 62.50 & 64.89 \\
        \bottomrule
    \end{tabular}}
\end{table}

We conduct the AndroidDaily evaluation on iOS devices. AndroidDaily contains many long-horizon tasks that require cross-application analysis and an explicit answer rather than a single visual goal. Table~\ref{tab:androiddaily-completion} shows that \ourmodel's largest margins over GUI-only baselines fall on \emph{Analyze} and \emph{complex} (Comp.) tasks, where explicit subtask decomposition, blackboard information transfer, and retrieved memory reduce repeated trajectory solving. The difference between the resolved and all settings reflects app availability and environment mismatch: these factors, rather than policy quality alone, account for a substantial fraction of end-to-end failure.

\FloatBarrier

\subsection{Case Study}
\label{sec:case-study}

This case study uses representative workflows to illustrate four characteristics of \ourmodel: routing and information transfer, experience-memory correction, navigation compression through skills and shortcuts, and host recovery and direct participation. The screenshots come from actual executions; host cards, step labels, per-step thoughts, and blackboard pills are overlays added for readability.

\begin{figure}[!ht]
    \centering
    \includegraphics[width=0.84\linewidth]{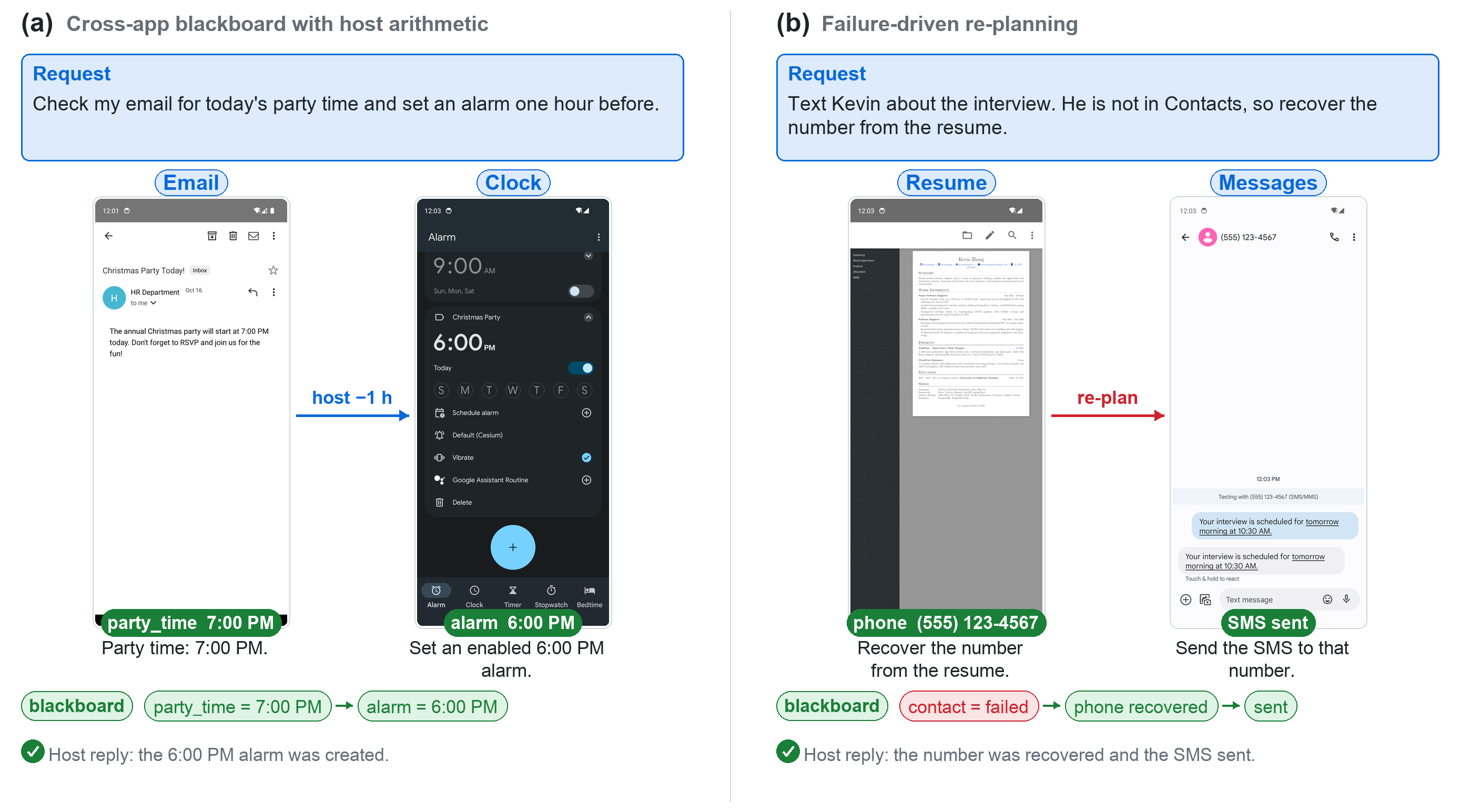}
    \caption{Cases of our workflow with attribution experience. \textbf{(a)}~Email-to-alarm transfers the observed party time through the blackboard; the host subtracts one hour before the GUI subagent sets the alarm. \textbf{(b)}~Failure-driven re-planning records a failed contact lookup, recovers the phone number from a resume, and delegates the final Messages GUI task.}
    \label{fig:case-workflow-patterns}
\end{figure}

\begin{figure}[!t]
    \centering
    \includegraphics[width=0.82\linewidth,height=0.54\textheight,keepaspectratio]{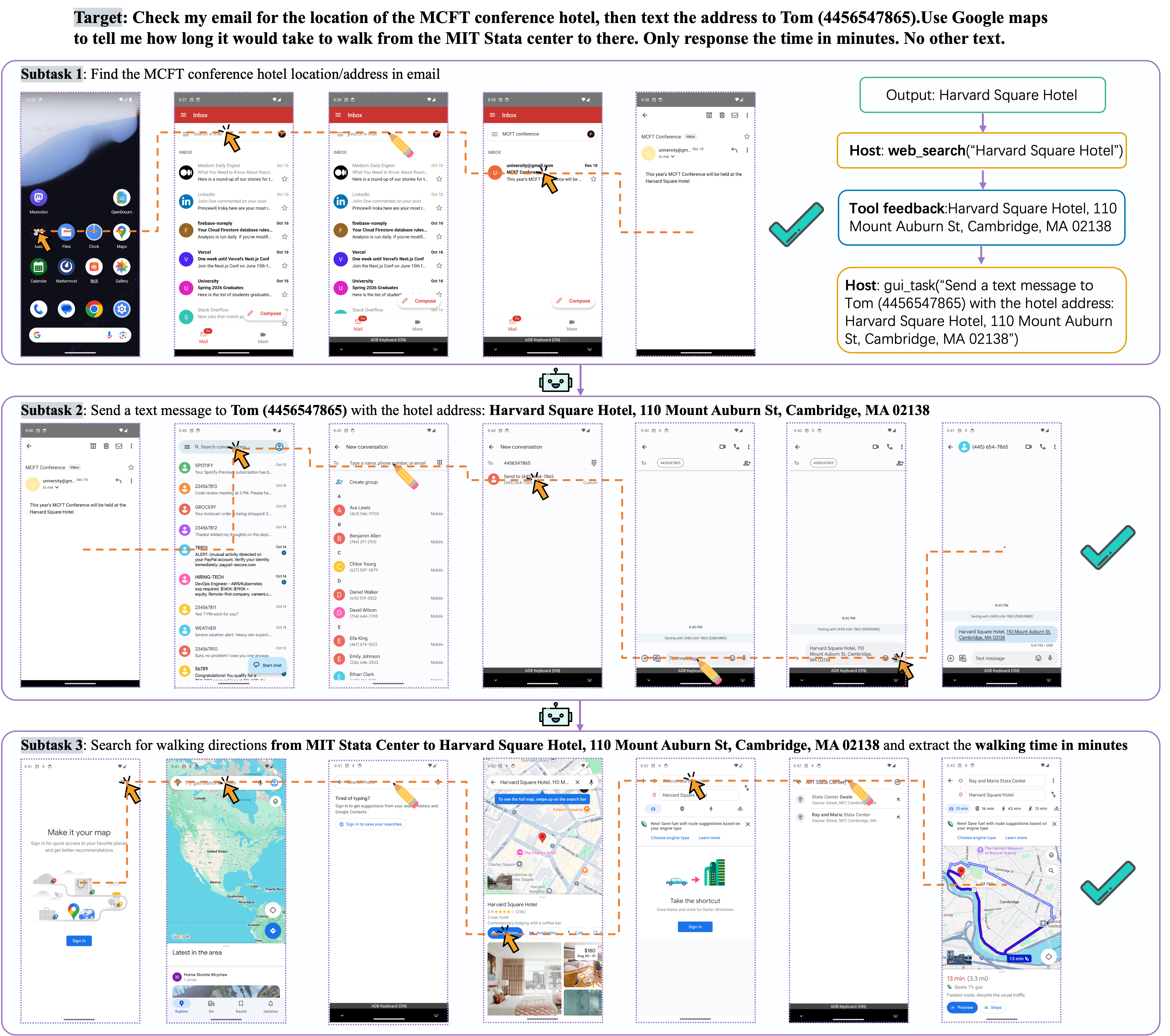}
    \caption{Host-mediated recovery in a conference-location task. The Email GUI task returns only the hotel name, so the host resolves the full address through web search before delegating the Messages and Maps GUI tasks. Maps reports a $13$-minute walk. The workflow combines partial GUI evidence with external tools while preserving typed subtask boundaries.}
    \label{fig:tool-gui-case}
\end{figure}

\noindent\textbf{Routing and information transfer.} Long-horizon, cross-app tasks amplify memory decay and goal drift when one GUI trajectory must retain every intermediate summary. \ourmodel instead decomposes the request into app-scoped GUI tasks and passes only declared outputs through the blackboard. Figure~\ref{fig:case-workflow-patterns}(a) isolates this handoff in an Email-to-Clock workflow, while Figure~\ref{fig:tool-gui-case} shows a longer Email-to-Messages-to-Maps workflow in which the resolved address becomes an explicit input to both downstream GUI tasks. The cart-to-SMS case in Appendix~\ref{app:supplementary-case-studies} similarly transfers product names, an order number, and a recipient phone number from shopping to messaging (Figure~\ref{fig:cart-sms-case}). These bounded handoffs reduce the history each GUI task must preserve and keep later goals tied to explicit inputs.

\noindent\textbf{Experience-memory correction.} Experience memory supplies prior failure lessons as advisory task context before low-level control begins. Figure~\ref{fig:gui-memory-case} contrasts an invite-link task with and without retrieved memory. Without memory, GUIClaw remains in Mastodon's mobile settings and follows a nonproductive path; with memory, it recalls that advanced invite links require the web administration panel and switches to the productive interface. The retrieved lesson corrects app and route selection without replacing live screen evidence.

\begin{figure}[!ht]
    \centering
    \includegraphics[width=0.98\linewidth]{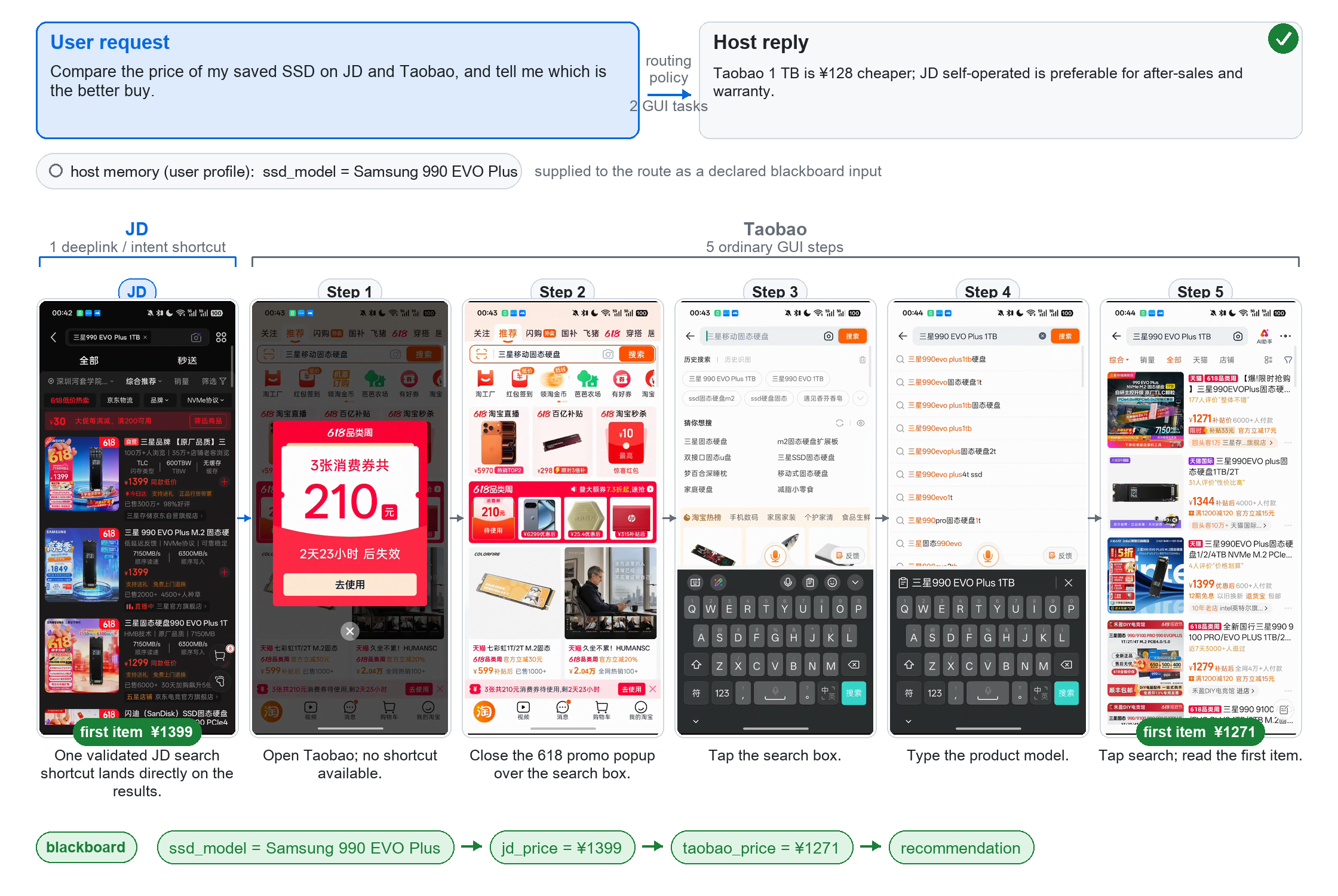}
    \caption{\textbf{KnowAct-GUIClaw execution of a cross-app price comparison}. The host supplies the product model from the user profile and routes two GUI tasks. A validated JD search shortcut lands directly on the results page, whereas Taobao requires five ordinary GUI steps. The blackboard carries both observed prices into the host's recommendation, illustrating the step and token savings in Table~\ref{tab:skill-reuse}.}
    \label{fig:case1-trajectory}
\end{figure}

\noindent\textbf{Navigation compression through skills and shortcuts.} Skills compress repeated procedures, while validated shortcuts eliminate navigation when they can reach a task-relevant page directly. In Figure~\ref{fig:case1-trajectory}, one validated JD search shortcut lands on the product-results page, whereas the corresponding Taobao branch requires five ordinary GUI steps. In Figure~\ref{fig:cart-sms-case}, a validated messaging shortcut opens the SMS compose view with the recipient phone number and message body already populated, leaving GUIClaw to verify and send the message. These cases visualize the navigation savings measured more broadly on skill-using tasks in Table~\ref{tab:skill-reuse}.

\noindent\textbf{Host recovery and direct participation.} The host intervenes when GUI execution fails or returns a value that does not satisfy a downstream contract. Figure~\ref{fig:case-workflow-patterns}(b) shows failure-driven re-planning: after a message-by-name attempt fails because the contact is absent, the host records the failed lookup, recovers the phone number from a resume, and delegates a new Messages GUI task. Figure~\ref{fig:tool-gui-case} shows direct tool participation after the Email GUI task returns only the ambiguous hotel name ``Harvard Square Hotel.'' The host resolves the full street address through web search before delegating the SMS and Maps GUI tasks, avoiding a longer GUI-only address lookup; the final Maps task reports a $13$-minute walk. The two trajectories distinguish host-side re-planning from tool-assisted information resolution while preserving typed subtask boundaries.

\begin{table}[t]
    \caption{MobileWorld repeated-attempt upper bound for the A--F configurations of Table~\ref{tab:efficiency}. ``Any of 3'' counts a task solved if at least one of three runs succeeds; ``all 3'' requires all three. SR (pass@1) is repeated for reference. This shows our model steadily boosts accuracy without large random volatility.}
    \label{tab:pass3}
    \centering
    \footnotesize
    \begin{tabular}{lcccccc}
        \toprule
        Metric & A & B & C & D & E & F \\
        \midrule
        single-run SR (\%)    & 24.8 & 34.5 & 37.9 & 40.7 & 43.3 & 46.2 \\
        pass@3, any of 3 (\%) & 34.2 & 47.9 & 51.3 & 51.3 & 59.0 & 59.8 \\
        pass@3, all 3 (\%)    & 15.4 & 18.8 & 26.5 & 30.8 & 29.1 & 32.5 \\
        \bottomrule
    \end{tabular}
\end{table}

\FloatBarrier

\subsection{Cross-Platform Checks}

We further test whether the same KnowAct-GUIClaw interface remains usable outside the Android evaluation stack. On HarmonyOS, we run MobileWorld-derived tasks that are not tied, or only weakly tied, to Android-specific mirrored app states; weakly tied cases are manually initialized before evaluation, and outcomes are judged with \texttt{qwen3.5-flash}. On Windows, we use a manually designed desktop set to check basic usability across browser, file, office, terminal, and system-control workflows. \textit{Under this protocol, \ourmodel solves $48/63$ HarmonyOS tasks ($76.2\%$) and $21/30$ Windows tasks ($70.0\%$).} Table~\ref{tab:cross-platform-cases} reports ten source-matched cases from each platform.

{\footnotesize
\setlength{\LTpre}{0.75\baselineskip}
\setlength{\LTpost}{0.75\baselineskip}
\setlength{\LTcapwidth}{\linewidth}
\begin{longtable}{@{}>{\raggedright\arraybackslash}p{0.89\linewidth}c@{}}
    \caption{Representative cross-platform usability cases. HarmonyOS instructions are translated into English when necessary while preserving their source requirements. Windows instructions name one concrete application wherever the source offered alternatives or referred to a generic application.}
    \label{tab:cross-platform-cases} \\
    \toprule
    Instruction & Success \\
    \midrule
    \endfirsthead
    \multicolumn{2}{c}{\tablename~\thetable\ continued} \\
    \toprule
    Instruction & Success \\
    \midrule
    \endhead
    \midrule
    \multicolumn{2}{r}{Continued on next page} \\
    \endfoot
    \bottomrule
    \endlastfoot
    \multicolumn{2}{@{}l}{\textit{HarmonyOS mobile checks}} \\
    Next Saturday from 10:00 a.m.\ to 12:30 p.m., I will travel to Shanghai Hongqiao Railway Station. Add a Calendar event named ``Business Trip.'' Find attractions within 10~km of the station that I can visit before work on Monday, and place them in the event description as ``attraction name: address,'' separated by commas. & Yes \\
    Check my calendar and send a WeChat message to Su with the dates of my arrival in Shanghai. The message must contain only the two dates in \texttt{MM/DD/YYYY} format, separated by a comma. & Yes \\
    I received a coffee invitation for 3:00 p.m.\ tomorrow. Check my calendar; if I am available, reply ``OK'' to Ping An Xi Le in QQ and create the corresponding calendar event. Otherwise, reply ``Not available in this time slot.'' & Yes \\
    Plan exactly one shortest taxi route in Chengdu from Chengdu Shuangliu International Airport Terminal~2 to my hotel at No.~8, South Section of Chunxi Road, Jinjiang District, visiting exactly Kuanzhai Alley and Jinli Ancient Street in either order. Send Ping An Xi Le in QQ the names and coordinates of all four locations as ``name: longitude, latitude,'' the shortest visit order, the three driving distances in meters, and the total driving distance in meters; separate each ordered list with commas. & Yes \\
    An error message on my phone contains the keyword \texttt{TurnOffWifi}. Search the \texttt{google-research/android\_world} GitHub repository for related issues. If similar issues exist, send every issue link, separated by commas, to Ping An Xi Le in QQ; otherwise send ``no turnoffwifi issues.'' & Yes \\
    Find the resume file downloaded most recently within the past month in Downloads, and send it to my HR colleague with the subject \texttt{candiaditaes\_cv}. & Yes \\
    Take a selfie and share it with Jimmy via email. & Yes \\
    Increase the font size and icons on my phone to the maximum setting. & Yes \\
    Set a weekend alarm for 8:25 a.m.\ with the ringtone ``beebeep'' and vibration off. & No \\
    Halloween is approaching. Use TaoBao to place an order for a set of temporary tattoos, and hand control back to me when the payment page appears. & No \\
    \hline
    \addlinespace
    \multicolumn{2}{@{}l}{\textit{Windows desktop checks}} \\
    Open the Ctrip website and search for a one-way flight from Beijing to Tokyo on the Friday after next. Filter for nonstop flights, sort by price in ascending order, and select the cheapest option. & Yes \\
    Open Baidu Maps in Google Chrome and search for ``Starbucks near Tiananmen Square.'' Select the highest-rated result from the left-hand list, open its details, copy its building or street address, and save the address to \texttt{address.txt} on the desktop using Notepad. & Yes \\
    Open \texttt{loop.py} in Visual Studio Code, add a breakpoint on line~3, open Run and Debug, and start debugging. & Yes \\
    Use Google Chrome to find the current Apple (AAPL) share price, calculate the value of 100 shares, and save the total to \texttt{stock.txt} on the desktop using Notepad. & Yes \\
    Open Windows Settings, retrieve the detailed operating-system version, and save it in a new draft email in Microsoft Outlook. & Yes \\
    Open the PDF report in Documents with Microsoft Edge, translate the first paragraph of Chapter~1 into English using Google Translate, and save the translation as a plain-text file on the desktop using Notepad. & Yes \\
    Open GitHub in Google Chrome, search for the \texttt{nanobot} project, copy its latest commit hash, and save it to \texttt{commit\_record.txt} on the desktop using Notepad. & Yes \\
    Open Microsoft PowerPoint, create a blank slide, insert a local image, and apply a ``Fly In'' animation to the image. & Yes \\
    I will send a test message in Slack within the next minute. Keep the desktop idle; when the Slack notification appears, click it immediately, open the conversation, and reply ``Received.'' & No \\
    Open the locally installed WeChat, open File Transfer Assistant, send the text ``Test emoji,'' and then use the built-in emoji panel to send a ``laughing'' emoji. & No \\
\end{longtable}
}

\noindent\textbf{HarmonyOS failure modes.} The failed HarmonyOS cases expose both low-level control difficulties and state-semantic mismatches. In the weekend-alarm case, the minute field uses an inertial wheel picker. GUIClaw repeatedly swipes within a nearby range but cannot settle on 25 minutes, so it exhausts the step budget without setting 8:25 a.m. In the temporary-tattoo case, the agent stops on the order-confirmation screen before submitting the order, rather than handing control back after reaching the subsequent payment page. This premature termination reflects application-level UI diversity: after a payment method is selected, TaoBao presents a ``Pay Now'' button, which the model misinterprets as evidence that the current screen is already the payment page; in fact, the payment page appears only after that button is pressed. The full set also shows system-UI mismatches: several quick-setting tasks repeatedly open the left notification pane even though the relevant HarmonyOS controls require a swipe from the right.

\noindent\textbf{Windows failure modes.} Windows failures expose three distinct grounding bottlenecks. First, the WeChat case reaches File Transfer Assistant and sends the text correctly, but selects a different face from the built-in panel instead of the requested laughing emoji. This error reveals a limitation in mapping a natural-language affect label to the corresponding visually similar, unlabeled icon. Second, the notification case requires sustained observation followed by an immediate click on a short-lived toast, exposing weak temporal grounding. The remaining failures primarily involve continuous geometric control, deep application-specific controls, and targets outside the current viewport, as seen in free-form canvas manipulation, window arrangement, and bottom-of-page actions. Thus, desktop reliability depends not only on pointer precision and timing, but also on semantic grounding between language and graphical symbols.

\subsection{Reproducibility}

MobileWorld results use the benchmark's deterministic evaluators on the 117 GUI-Only tasks at a 50-step cap. We log GUI-executor traces and host calls separately, so that the GUI-trace total and host total columns of Table~\ref{tab:efficiency} can be reproduced independently. AndroidDaily scoring (\texttt{qwen3.5-flash} for GUI-only tasks, two human experts under the $1.0/0.5/0$ rule for answer-returning tasks) and the resolved/all splits follow the protocol described above. You can see the experimental logs at \url{https://github.com/HITsz-TMG/KnowAct/releases/tag/Result}.

\section{Conclusion}
\label{sec:conclusion}
This work proposes KnowAct-GUIClaw, a cross-platform GUI agent framework built on the ``Know Deeply, Act Perfectly" paradigm to address two core limitations of mainstream OpenClaw-style agent systems: insufficient cross-device GUI interaction capacity and the absence of native self-evolution mechanisms. The framework instantiates a four-stage Know–Route–Act–Reflect closed-loop pipeline that decouples high-level task orchestration (host agent) from low-level visual device manipulation (GUI subagent). It introduces two core reusable storage modules: attribution-aware persistent experience memory and state-validated self-evolving skill libraries, alongside a typed blackboard information transfer protocol to standardize cross-application data flow and eliminate fuzzy free-text context loss. Future work targets tighter native integration of the Knowledge module, external general-purpose tools, and the GUI subagent to eliminate rigid sequential pipeline handoffs. We will design a unified joint planner to concurrently weigh knowledge retrieval, non-visual tool calls and GUI actions within one decision cycle and cut host-subagent communication overhead.

\section{Contributors}
\label{sec:contributors}

\textbf{Core Contributors} 

Yunxin Li, Jinchao Li, Baotian Hu, Min Zhang

\textbf{Contributors}

Shibo Su, Zhenran Xu, Chenrui Zhao, Tongshu Bian, Xiaoman Liang, Meishan Zhang

\textbf{Corresponding Author}

Baotian Hu

Harbin Institute of Technology, Shenzhen; Shenzhen Loop Area Institute

Email: hubaotian@hit.edu.cn


\bibliographystyle{lychee}
\bibliography{custom}

\appendix
\section{Supplementary Case Studies}
\label{app:supplementary-case-studies}
\label{app:case-study-annotation}

\begin{figure}[p]
    \centering
    \includegraphics[width=\linewidth]{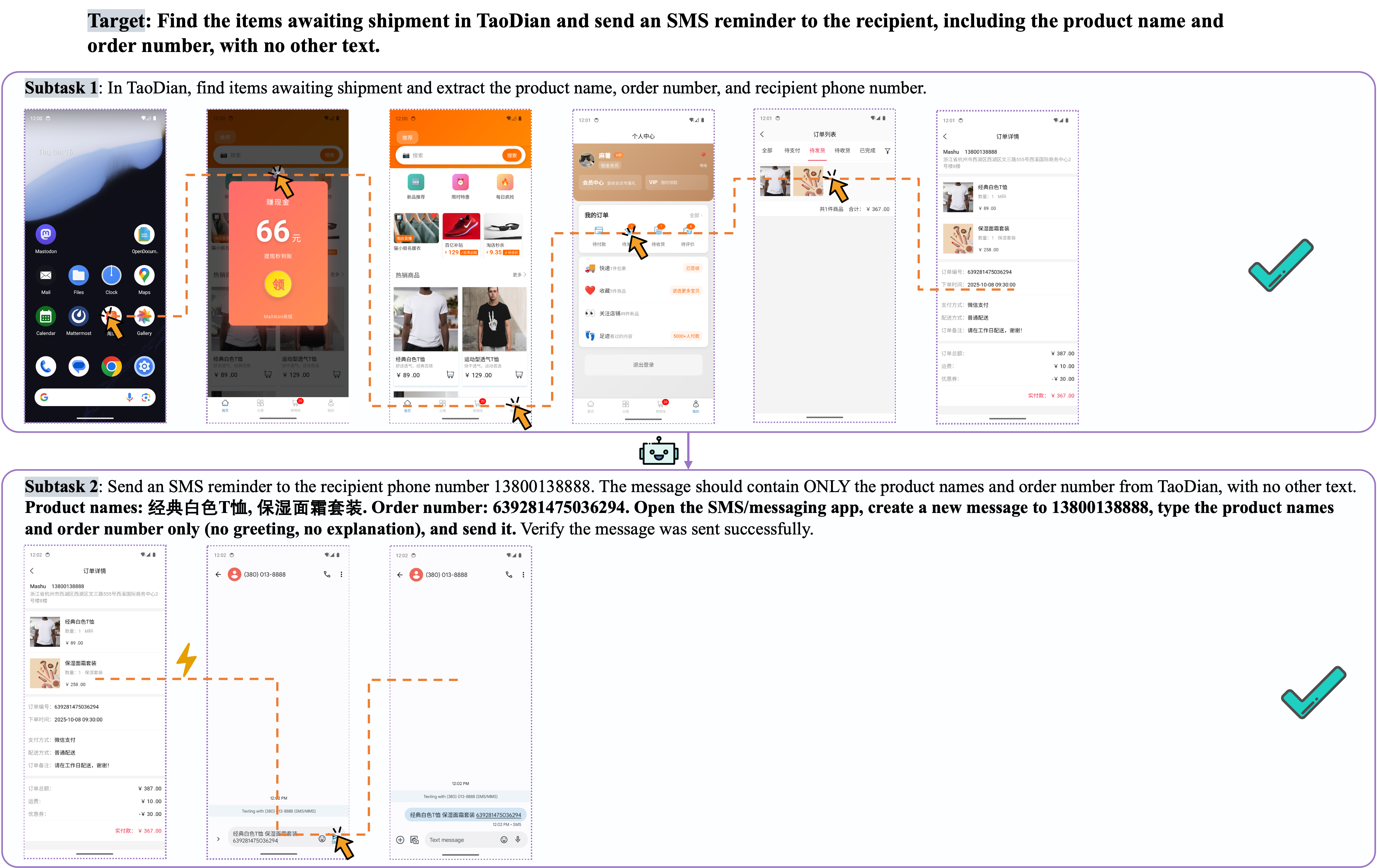}
    \caption{Cart-to-SMS cross-app execution. The TaoDian GUI task extracts the product names, order number, and recipient phone number and transfers them to the downstream messaging task. A validated messaging shortcut opens the SMS compose view with the recipient and message body already populated, illustrating both blackboard information transfer and navigation compression.}
    \label{fig:cart-sms-case}
\end{figure}

Figure~\ref{fig:cart-sms-case} supplements the routing and navigation-compression cases in Section~\ref{sec:case-study}. The first GUI task extracts the product names, order number, and recipient phone number from TaoDian as explicit outputs. A validated messaging shortcut then opens the SMS compose view with the recipient and message body populated, leaving GUIClaw to verify and send the message.

\begin{figure}[p]
    \centering
    \includegraphics[width=\linewidth,height=0.70\textheight,keepaspectratio]{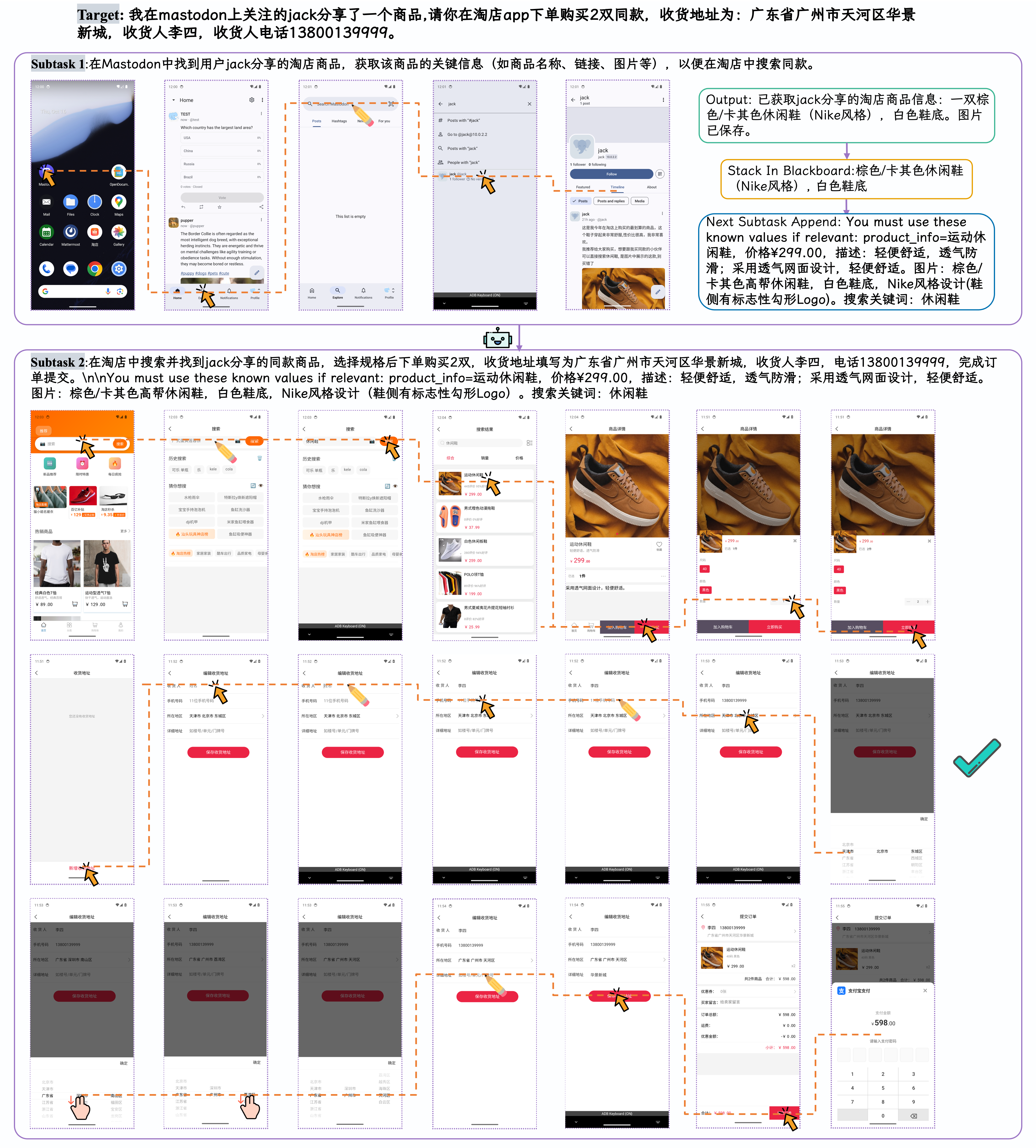}
    \caption{Chinese annotated companion for the host-mediated recovery case in Figure~\ref{fig:tool-gui-case}. The panel highlights the same control flow in which an underspecified GUI output is routed back to the host, grounded through web search, and then consumed by downstream Messages and Maps GUI tasks.}
    \label{fig:tool-gui-case-cn}
\end{figure}

Figure~\ref{fig:tool-gui-case-cn} provides an annotated Chinese companion for the host-mediated recovery case in Figure~\ref{fig:tool-gui-case}. It shows the same control flow in which an underspecified GUI output is routed back to the host, grounded through web search, and then consumed by downstream Messages and Maps GUI tasks.

\section{Action Space}
\label{app:action-space}

Table~\ref{tab:action-space} gives the unified action space GUIClaw exposes to the GUI executor, realizing $\mathcal{A}_{\mathrm{gui}}\cup\mathcal{A}_{\mathrm{ask}}$ from Section~\ref{sec:method_skills}. The two platforms share a common core of pointer, text, navigation, and task-control actions; desktop adds key combinations, an explicit pointer swipe, and application launch and close, while mobile adds a dedicated enter key. This largely shared action surface is what lets a single host drive both platforms through the same GUI task interface.

\begin{table}[H]
    \caption{Unified action space of GUIClaw across platforms. \textbf{Shared} actions are available on both mobile and desktop; \textbf{Desktop} and \textbf{Mobile} list the platform-specific additions. Together they realize $\mathcal{A}_{\mathrm{gui}}\cup\mathcal{A}_{\mathrm{ask}}$ (Section~\ref{sec:method_skills}); skills ($\mathcal{A}_{\mathrm{skill}}$), Android deeplink/intent shortcuts ($\mathcal{A}_{\mathrm{shortcut}}$), and external Model Context Protocol (MCP) tool calls extend this set through the runtime.}
    \label{tab:action-space}
    \centering
    \small
    \begin{tabular}{l l l}
        \toprule
        \textbf{Environment} & \textbf{Action} & \textbf{Definition} \\
        \midrule
        \multirow{11}{*}{\textbf{Shared}}
          & Click($x, y$)                & Clicks or taps at coordinates $(x, y)$. \\
          & DoubleTap($x, y$)            & Double-clicks or double-taps at $(x, y)$. \\
          & LongPress($x, y$)            & Long-presses at $(x, y)$. \\
          & Drag($x_1, y_1, x_2, y_2$)   & Drags from $(x_1, y_1)$ to $(x_2, y_2)$. \\
          & Scroll($x, y$, direction)    & Scrolls at $(x, y)$ in the given direction. \\
          & Type(content)                & Types the specified content. \\
          & Back()                       & Returns to the previous screen. \\
          & Home()                       & Returns to the home screen. \\
          & Wait()                       & Pauses for a brief moment. \\
          & Finished()                   & Returns the final answer or marks the task complete or infeasible. \\
          & CallUser()                   & Requests user intervention. \\
        \midrule
        \multirow{4}{*}{\textbf{Desktop}}
          & Hotkey(key)                  & Presses the specified key combination. \\
          & Swipe($x_1, y_1, x_2, y_2$)  & Swipes the pointer from $(x_1, y_1)$ to $(x_2, y_2)$. \\
          & OpenApp(name)                & Launches the specified application. \\
          & CloseApp(name)               & Closes the specified application. \\
        \midrule
        \textbf{Mobile}
          & PressEnter()                 & Presses the ``enter'' key. \\
        \bottomrule
    \end{tabular}
\end{table}

\section{Skill Extraction and Prompt Contracts}
\label{app:skill-induction}

\ourmodel uses prompt contracts at both execution time and reflection time. The execution prompt below is the runtime MobileWorld-style template used for prompt-side skill selection~\citep{kong2025mobileworldbenchmarkingautonomousmobile}: GUIClaw retrieves top-$k$ relevant skills, formats them as the \texttt{Compact skills} catalog, and lets the GUI executor choose a listed skill or continue with ordinary actions. Reflection then turns GUI experience into reusable knowledge through additional contracts rather than unrestricted trace replay. First, trajectory evidence is normalized into a task, an app scope, ordered actions, screenshots, target hints, and state contracts. Second, a skill-extraction agent compresses this evidence into one reusable, parameterized procedure and rejects outputs that are not executable or not state guarded. Third, a separate evolution agent repairs a previously reused skill only when a failure trace identifies that skill as the failure source. Finally, a memory-induction agent writes textual success and failure lessons that guide future routing and execution without becoming executable actions. The boxes below reproduce the runtime source prompts without shortening; braces denote runtime placeholders and structured inputs supplied by the caller.

\tcbset{
    skillbox/.style={
        enhanced, breakable,
        listing only,
        listing options={
            language=Python,
            basicstyle=\ttfamily\scriptsize,
            keywordstyle=\bfseries\color{academicblue!85!black},
            commentstyle=\itshape\color{gray!55!black},
            stringstyle=\color{OliveGreen!70!black},
            morekeywords={async,await},
            alsoletter={_},
            emph={click_then_type,search_product,jd_splash,jd_search},
            emphstyle=\bfseries\color{teal!70!black},
            literate={@skill}{{\textcolor{RoyalPurple!85!black}{\bfseries @skill}}}6,
            showstringspaces=false,
            columns=fullflexible,
            keepspaces=true,
            breaklines=true,
            breakatwhitespace=false,
        },
        colback=gray!2,
        colframe=academicblue!55,
        coltitle=black,
        colbacktitle=academicblue!12,
        fonttitle=\bfseries\small,
        boxrule=0.5pt,
        arc=2pt,
        left=5pt, right=4pt, top=3pt, bottom=3pt, boxsep=2pt,
    },
    promptbox/.style={
        enhanced, breakable,
        listing only,
        listing options={
            basicstyle=\ttfamily\scriptsize,
            keywordstyle=\bfseries,
            commentstyle=\itshape\color{gray!55!black},
            stringstyle=\color{gray!40!black},
            showstringspaces=false,
            columns=fullflexible,
            keepspaces=true,
            breaklines=true,
            breakatwhitespace=false,
        },
        coltitle=black,
        fonttitle=\bfseries\small,
        boxrule=0.5pt,
        arc=2pt,
        left=5pt, right=4pt, top=3pt, bottom=3pt, boxsep=2pt,
    },
    reusepromptbox/.style={
        promptbox,
        colback=gray!3,
        colframe=gray!60,
        colbacktitle=gray!16,
    },
    execpromptbox/.style={
        promptbox,
        colback=SkyBlue!4,
        colframe=SkyBlue!65,
        colbacktitle=SkyBlue!14,
    },
    extractpromptbox/.style={
        promptbox,
        colback=academicblue!4,
        colframe=academicblue!65,
        colbacktitle=academicblue!14,
    },
    evolvepromptbox/.style={
        promptbox,
        colback=OliveGreen!5,
        colframe=OliveGreen!65,
        colbacktitle=OliveGreen!15,
    },
    shortcutpromptbox/.style={
        promptbox,
        colback=RoyalPurple!4,
        colframe=RoyalPurple!60,
        colbacktitle=RoyalPurple!14,
    },
    memorypromptbox/.style={
        promptbox,
        colback=BurntOrange!5,
        colframe=BurntOrange!65,
        colbacktitle=BurntOrange!16,
    },
}

\begin{tcblisting}{execpromptbox, title={(a)~Execution Prompt}}
# Role: Android Phone Operator AI
You are an AI that controls an Android phone to complete user requests. Your responsibilities:
- Answer questions by retrieving information from the phone.
- Perform tasks by executing precise actions.

# Action Framework
Respond with EXACT JSON format for one of these actions:
| Action          | Description                              | JSON Format Example                                                         |
|-----------------|----------------------------------------- |-----------------------------------------------------------------------------|
| `click`         | Tap visible element (describe clearly)   | `{"action_type": "click", "coordinate": [x, y]}`   |
| `double_tap`    | Double-tap visible element (describe clearly)   | `{"action_type": "double_tap", "coordinate": [x, y]}`   |
| `long_press`    | Long-press visible element (describe clearly) | `{"action_type": "long_press", "coordinate": [x, y]}`            |
| `drag`          | Drag from visible element to another visible element (describe both clearly) | `{"action_type": "drag", "start_coordinate": [x1, y1], "end_coordinate": [x2, y2]}`            |
| `input_text`    | Type into field | `{"action_type":"input_text", "text":"Hello"}|
| `answer`        | Respond to user                          | `{"action_type":"answer", "text":"It's 25 degrees today."}`               |
| `navigate_home` | Return to home screen                    | `{"action_type": "navigate_home"}`                                        |
| `navigate_back` | Navigate back                            | `{"action_type": "navigate_back"}`                                        |
| `scroll`        | Scroll direction (up/down/left/right)    | `{"action_type":"scroll", "direction":"down"}`                            |
| `status`        | Mark task as `complete` or `infeasible`  | `{"action_type":"status", "goal_status":"complete"}`                      |
| `wait`          | Wait for screen to update                | `{"action_type":"wait"}`                                                  |
| `ask_user`      | Ask user for information                 | `{"action_type":"ask_user", "text":"what is the exact requirements do you need?"}`        |
| `keyboard_enter`   | Press enter key         | `{"action_type":"keyboard_enter"}`               |
| `use_skill`     | Run a listed compact GUI skill prefix when it clearly matches the task | `{"action_type":"use_skill","skill_id":"listed_skill_id","arguments":{}}` |
| `click_then_type` | Preferred one-step action for visible text fields: tap a coordinate and type text. Use auto_enter true only for search submission. | `{"action_type":"click_then_type","coordinate":[x,y],"text":"Hello","auto_enter":false}` |
| `click_multi` | Tap multiple visible coordinates in sequence | `{"action_type":"click_multi","coordinates":[[x1,y1],[x2,y2]]}` |

Note:
- The coordinate is the center of the element to be clicked/long-pressed/dragged.
- x, y are coordinates in the screen, the origin is the top-left corner of the screen.
- x, y are numbers, the range is normalized to [0, 1000].

# Execution Principles
1. Communication Rule:
   - ALWAYS use 'answer' action to reply to users - never assume on-screen text is sufficient
   - Please follow the user instruction strictly to answer the question, e.g., only return a single number, only return True/False, only return items separated by comma.
   - NEVER use 'answer' action to indicate waiting or loading - use 'wait' action instead
   - Note that `answer` will terminate the task immediately.

2. Efficiency First:
   - Choose simplest path to complete tasks
   - If action fails twice, try alternatives (e.g., long_press instead of click)

3. Smart Navigation:
   - Gather information when needed (e.g., open Calendar to check schedule)
   - For scrolling:
     * Scroll direction is INVERSE to swipe (scroll down to see lower content)
     * If scroll fails, try opposite direction

4. Text Operations:
   - You MUST first click the input box to activate it before typing the text.
   - When a text field is visible, prefer `click_then_type` (one action) over separate `click`+`input_text` to activate and type known text.
   - For text manipulation:
     1. Long-press to select
     2. Use selection bar options (Copy/Paste/Select All)
     3. Delete by selecting then cutting

5. Ask User:
    - If you think you have no enough information to complete the task, you should use `ask_user` action to ask the user to get more information.


# Decision Process
0. Before a manual GUI action, check the compact skill list; if one clearly matches the requested app/workflow, choose `use_skill` (it may open the target app internally, so do not open it manually first).
0a. Use listed composite actions only when they exactly match the next local operation. Prefer `click_multi` when several targets on screen need the same tap and none opens a confirmation dialog. Prefer `click_then_type` over separate `click`+`input_text` when tapping a visible text field to enter known text -- unless the field is already focused, existing text must be cleared, tapping opens a picker to observe first, or the text depends on the tap result.
1. Analyze goal, history, and current screen
2. Determine if task is already complete (use `status` if true)
3. If not, choose the most appropriate action to complete the task.
4. Output in exact format below, and ensure the Action is a valid JSON string:
5. The action output format is different for GUI actions and MCP tool actions. Note only one tool call is allowed in one action.

# Optional Compact GUI Skills
Optionally pick ONE listed compact skill as a single action when it clearly matches
the task. If none clearly matches, keep using the normal GUI actions above.

Skill action format:
`{"action_type":"use_skill","skill_id":"listed_skill_id","arguments":{"param":"value"}}`

Rules:
- Prefer `use_skill` over manual navigation when a listed skill clearly matches the
  requested app/workflow; copy its `skill_id` exactly.
- A skill may open/navigate the target app internally, so the target app need not
  already be on screen.
- Fill `arguments` only with values the task makes obvious; otherwise use `{}`.

Compact skills:
- skill_id={skill_id}; skill_name={skill_name}; description={description}; app={app}; parameters={comma_separated_parameters}

# Expected Output Format (`Thought: ` and `Action: ` are required):
Thought: [Analysis including reference to key steps/points when applicable]
Action: [Single JSON action]

# Output Format Example
## for GUI actions:
Thought: I need to ... to complete the task.
Action: {"action_type": "type", "text": "What is weather like in San Francisco today?"}
\end{tcblisting}

\begin{tcblisting}{extractpromptbox, title={(b)~Skill extraction prompt}}
Extract a reusable GUI skill as Python code from this trajectory.

Target format:
from opengui.skills.flat import C, R, action, skill, tag

@skill(app="{app}", platform="{platform}", name="short_name", description="One sentence summary")
async def skill_name(device, param1, param2):
    await action("open_app", target="{app}", fixed=True, fixed_values={"text": "{app}"},
                 valid_state="No need to verify")
    await action("tap", target="search button", fixed=True,
                 fixed_values={"x": 540, "y": 960},
                 valid_state="target is visible and clickable",
                 state_contract=C(app="...[app_package]...", required=[R(resource_id="target_id", visible=True)]))
    await action("input_text", target="{{param1}}",
                 valid_state="input field is focused")

Rules:
- Extract ONE cohesive skill that covers the core action sequence. Do NOT split into multiple tiny @skill functions.
- fixed=true + fixed_values: static UI (nav bars, toolbar, system actions, open_app). Copy exact x/y/text from the trajectory step params shown after "|".
- Use only these action types: tap, long_press, double_tap, drag, swipe, scroll, input_text, hotkey, screenshot, wait, open_app, open_deeplink, open_intent, close_app, back, home, enter, app_switch, done, request_intervention. Do not invent actions such as read_text, press_key, navigate_back, or navigate_to_folder.
- The skill function body must contain only await action(...) statements. Do not add if/for/while/try blocks, assignments, calculations, helper calls, return values, or non-action awaits.
- Do not use f-strings, string concatenation, arithmetic expressions, comparisons, comprehensions, dict/list expressions with computed values, or nested function calls in action arguments. Use literal strings or {{param}} placeholders only.
- fixed_values may contain only executable action fields such as x, y, x2, y2, text, key, pixels, direction, duration_ms, component, package, intent_action, mime_type, categories, extras, relative, status, auto_enter. Never put selectors such as resource_id, content_desc, class, or class_name in fixed_values.
- fixed=false: dynamic content (search results, variable input). Use {{param}} placeholders, omit fixed_values.
- target: Every required interactive step must have a natural-language target and valid_state. Use a concise natural-language grounding hint, e.g. "search button", "search input field", "matching video result", "skip ad button". Do not use raw class/resource_id as target unless it is also visible user-facing text.
- Collapse all app-launch steps into ONE open_app as the first step. For open_app, prefer the trajectory app package for both target and fixed_values.text when available, and always use valid_state="No need to verify".
- valid_state: Every required interactive step must have a specific present-tense valid_state, e.g. "search field is visible and enabled". For input_text, use "input field is focused" if no better state is available. If a required step has no verifiable state, remove or regenerate that step instead of leaving valid_state empty.
- state_contract: do not invent selectors. Copy only the exact contract provided by trajectory/codegen for the matched step; omit if no contract is provided. The extractor postprocess will align contracts from codegen.
- R(...) supports resource_id, text, content_desc, class_, xpath, visible, clickable, enabled, focused, and scrollable only. Do not use class_name.
- Drop duplicate/redundant clicks, exploratory taps, and pointless scrolls.
- Transient popups (ads, permissions, consent): keep as optional=True step. Executor skips them when absent.
- Keep transient blockers and benign app confirmations such as skip/close/save/done as guarded optional=True steps when they appear. Omit destructive or externally visible confirmations such as pay, delete, send, submit order, publish, or irreversible consent unless the original user task explicitly requires them.
- description: MUST be generic and reusable. Mention app name, capability, and broad feature-level route only. Use parameter roles like query, media item, contact, or item. NEVER include literal values/entities, exact titles/names, or narrow qualifiers such as specific, official, first result, or top result. Avoid tap-by-tap UI actions.

For failure trajectories, insert:
FAILURE trajectory: keep the reusable succeeded prefix. If the failure screen clearly shows one safe corrective next action, append at most one non-fixed corrective step with natural-language target and valid_state. Do not invent coordinates or state_contract for that corrective step. Use optional=True only for transient blockers/popups, and never add pay/delete/send/submit/publish/irreversible confirmation actions.

## Trajectory
{code_text}

Output ONLY the Python code. No markdown fences, no JSON object, no explanation.
\end{tcblisting}

\begin{tcblisting}{evolvepromptbox, title={(c)~Skill evolution prompt}}
You are improving one existing GUI automation skill after it failed during reuse.

Task:
{task}

Original skill JSON:
{skill_json}

Failure case:
{failure_case_json}

Prior feedback for this skill:
{feedback_json}

Full trajectory:
{trajectory_json}

Return ONLY a JSON object for the improved skill, using the same schema as the original skill.
Rules:
- Improve the original skill in place. Do not create a different skill or unrelated workflow.
- Keep the same high-level intent unless the failure shows the description caused a wrong match; then narrow the description/preconditions.
- If a popup or optional obstacle appeared, add a guarded optional step before the blocked step:
  {"action_type": "tap", "target": "Close", "parameters": {"optional": true}, "valid_state": "popup close button is visible"}
  Optional steps are skipped when their valid_state is not present.
- If an action target, parameter, valid_state, or state_contract is stale for the new UI, update that step.
- Preserve reusable parameters as {param_name} placeholders.
- Do not include final destructive actions such as pay, delete, send, submit, or irreversible confirmation unless the original skill already required them.
- Omit skill_id; the caller will preserve the original skill_id.
\end{tcblisting}

\begin{tcblisting}{shortcutpromptbox, title={(d)~Shortcut validation prompt}}
Verifier system prompt:
You validate Android deeplink/intent probe results. Return only one JSON object with fields: usable (boolean), status ('page_validated' or 'launchable' or 'failed'), name (short English function name, e.g. 'bili_search', 'taobao_cart'; required when usable=true), description (short natural-language capability, e.g. 'Bili video search'), parameters (array of strings), payload_preserved (boolean), reason (short string), next_variant_hint (string or null). Set usable=true only if the screenshot/UI indicates the intended app page opened and the payload, such as query text, was preserved when relevant. When usable=true for the intended page, set status='page_validated'; use status='launchable' only for target-package launches whose page purpose is unclear. The `name` field must be a concise English identifier (snake_case, max 30 chars) derived from the app and page function, e.g. 'bili_scan', 'taobao_search'. Do not treat a query visible only in search history or suggestions as payload_preserved; payload_preserved means the current input/result page reflects that payload. Use a concise natural-language description such as 'Bili video search'. Do not mention adb, URI, component, or implementation details in description or name.

Verifier user payload:
{
  "task": "{task}",
  "candidate": "{candidate_to_dict(candidate)}",
  "variant": "{variant_to_dict(variant)}",
  "foreground": "{foreground_app}",
  "adb_output": "{summarized_adb_output}",
  "ui_sample": ["..."],
  "ui_tree_excerpt": "{summarized_ui_tree}"
}
If available, the screenshot is attached as an image_url message part.

Metadata refinement system prompt:
You refine shortcut skill names/descriptions. Return only JSON.

Metadata refinement user prompt:
{
  "task": "Refine Android shortcut skill metadata after validation. Return JSON only: {\"records\":[{\"index\":0,\"name\":\"...\",\"description\":\"...\"}]}. Do not change payload fields, packages, status, parameters, valid_state, or indices. All promoted shortcut records must keep valid_state='No need to verify'. Prefer concise distinct names and descriptions that help an agent choose the right shortcut.",
  "records": "{compact_records}"
}
\end{tcblisting}

\begin{tcblisting}{memorypromptbox, title={(e)~Experience-memory induction prompt}}
Success system prompt:
You are an expert in Android GUI automation. You will be given a user task query
and the corresponding trajectory that represents **how an agent successfully
accomplished the task**.

## Guidelines
You need to extract and summarize useful insights in the format of memory items
based on the agent's successful trajectory.  The goal of summarized memory items
is to be helpful and generalizable for future similar tasks.

## Important notes
- You must first think why the trajectory is successful, and then summarize
  the insights.
- You can extract *at most 3* memory items from the trajectory.
- You must not repeat similar or overlapping items.
- Prefer concrete, actionable procedures over abstract principles.  Do not
  embed specific product names, queries, or literal string contents from the
  task.
- For Android GUI tasks, focus on: app navigation patterns, form-filling
  strategies, common UI pitfalls that were avoided, and efficient action
  sequences.
- When UI snippets are present in the trajectory, ground lessons in the actual
  visible/clickable controls rather than only repeating the agent's own
  reasoning.

## Output Format
Your output must strictly follow the Markdown format shown below:

```
# Memory Item i
## Title <the title of the memory item>
## Description <one sentence summary describing when to use the memory item>
## Content <1-3 sentences describing the insights learned to successfully
  accomplish similar tasks in the future>
```

Failure system prompt:
You are an expert in Android GUI automation. You will be given a user task query
and the corresponding trajectory that represents **how an agent attempted to
resolve the task but failed**.

## Guidelines
You need to extract and summarize useful insights in the format of memory items
based on the agent's failed trajectory.  The goal of summarized memory items is
to be helpful and generalizable for future similar tasks.

## Important notes
- You must first reflect and think why the trajectory failed, and then
  summarize what lessons you have learned or strategies to prevent the failure
  in the future.
- You can extract *at most 3* memory items from the trajectory.
- You must not repeat similar or overlapping items.
- Prefer concrete, actionable recovery procedures over abstract principles.
  Do not embed specific product names, queries, or literal string contents
  from the task.
- For Android GUI tasks, focus on: navigation mistakes, form-filling errors,
  premature task completion, app state assumptions that were wrong, and
  specific UI patterns that caused trouble.
- When UI snippets are present in the trajectory, ground lessons in the actual
  visible/clickable controls rather than only repeating the failed agent's own
  reasoning.

## Output Format
Your output must strictly follow the Markdown format shown below:

```
# Memory Item i
## Title <the title of the memory item>
## Description <one sentence summary describing when NOT to use this approach>
## Content <1-3 sentences describing the insights learned to avoid such
  failures in the future>
```

User prompt:
{trajectory_text}
\end{tcblisting}

\section{Skill and Shortcut Examples}
\label{app:skills}

The three boxes below show excerpts from \ourmodel's Android skill and shortcut store, abbreviated for readability: timestamps and some keyword arguments are dropped, 64-character state-contract hashes and long component names are truncated, and one Chinese accessibility label is glossed in ASCII. They illustrate the three forms covered by the single \emph{skill} abstraction (Section~\ref{sec:method_skills}): (a)~a parameterized click-then-type pattern in $\mathcal{A}_{\mathrm{skill}}$; (b)~a multi-step skill whose steps carry an expected state and, where available, a structural state contract checked before each step; and (c)~validated deeplink and intent shortcuts in $\mathcal{A}_{\mathrm{shortcut}}$. Stable values, such as package names, button coordinates, and intent components, are fixed in advance, while request-dependent values, such as the query string, are supplied at run time. The \texttt{validated} tag marks shortcut candidates that passed on-device validation rather than static manifest discovery alone.

\begin{tcblisting}{skillbox, title={(a)~Parameterized pattern: click-then-type}}
@skill(name='click_then_type', platform='android', tags=['compact_action'],
       skill_id='compact:action:click_then_type',
       description='Tap a [0,999] coordinate, then type text.')
async def click_then_type(device, coordinate, text):
    await action('click_then_type', target=coordinate, text=text,
                 auto_enter=False, valid_state='No need to verify')
\end{tcblisting}

\begin{tcblisting}{skillbox, title={(b)~Multi-step skill with per-step state validation (Pinduoduo)}}
@skill(name='search_product', app='com.xunmeng.pinduoduo', platform='android',
       skill_id='flat:search_product',
       description='Search for a product in Pinduoduo.')
async def search_product(device, query):
    await action('open_app', target='com.xunmeng.pinduoduo',
                 valid_state='No need to verify')
    await action('tap', target='search bar', fixed_values={'x': 242, 'y': 88},
                 valid_state='search bar is visible and clickable',
                 state_contract=C.from_dict({
                     'anchor': {'app_package': 'com.xunmeng.pinduoduo'},
                     'signature': {'required': [{
                         'selector': {'content_desc': 'search'},  # ASCII gloss
                         'state': ['visible', 'enabled']}]},
                     'fingerprint': '60c6a0fb...c950947b'}))  # sha256, truncated
    await action('input_text', target=query,
                 valid_state='input field is focused',
                 state_contract=C.from_dict({
                     'anchor': {'app_package': 'com.xunmeng.pinduoduo'},
                     'signature': {'required': [{
                         'selector': {'class': 'android.widget.EditText',
                                      'resource_id': 'com.xunmeng.pinduoduo:id/pdd'},
                         'state': ['visible', 'enabled', 'focused']}]},
                     'fingerprint': 'cbd6ef8c...4e7e2861'}))  # sha256, truncated
    await action('tap', target='search button', fixed_values={'x': 446, 'y': 96},
                 valid_state='search button is visible and clickable')
\end{tcblisting}

\begin{tcblisting}{skillbox, title={(c)~Validated deeplink and intent shortcuts (JD)}}
@skill(name='jd_splash', app='com.jingdong.app.mall', platform='android',
       tags=['shortcut', 'deeplink', 'validated'], success_count=1)
async def jd_splash(device):
    await action('open_deeplink', target='JDAnalytics:',
                 package='com.jingdong.app.mall', valid_state='No need to verify')

@skill(name='jd_search', app='com.jingdong.app.mall', platform='android',
       tags=['shortcut', 'intent', 'validated'], success_count=1)
async def jd_search(device, query):
    await action('open_intent', intent_action='android.intent.action.SEND',
                 package='com.jingdong.app.mall',
                 component='com.jingdong.app.mall/...SearchBridgeActivity',
                 mime_type='text/plain',
                 extras=[['android.intent.extra.TEXT', '{{query}}']],
                 valid_state='No need to verify')
\end{tcblisting}

These entries populate the skill and shortcut store that \emph{Know} retrieves and \emph{Act} applies during GUI task execution.

\end{document}